\documentclass[sigconf,authorversion,nonacm]{acmart}
\AtBeginDocument{%
  }

\renewcommand\footnotetextcopyrightpermission[1]{} %
\usepackage{graphicx}
\usepackage{subcaption}
\usepackage{tikz}
\usepackage{caption}
\usepackage{colortbl}
\usepackage{adjustbox}
\begin{document}

\title{3DGabSplat: 3D Gabor Splatting for Frequency-adaptive Radiance Field Rendering}

\author{Junyu Zhou$^\ast$}
\affiliation{%
  \institution{Shanghai Jiao Tong University}
  \city{Shanghai}
  \country{China}}
\email{blabla@sjtu.edu.cn}

\author{Yuyang Huang$^\ast$}
\affiliation{%
  \institution{Shanghai Jiao Tong University}
  \city{Shanghai}
  \country{China}}
\email{huangyuyang@sjtu.edu.cn}

\author{Wenrui Dai$^\dagger$}
\affiliation{%
  \institution{Shanghai Jiao Tong University}
  \city{Shanghai}
  \country{China}}
\email{daiwenrui@sjtu.edu.cn}

\author{Junni Zou$^\dagger$}
\affiliation{%
  \institution{Shanghai Jiao Tong University}
  \city{Shanghai}
  \country{China}}
\email{zoujunni@sjtu.edu.cn}

\author{Ziyang Zheng}
\affiliation{%
  \institution{Shanghai Jiao Tong University}
  \city{Shanghai}
  \country{China}}
\email{zhengziyang@sjtu.edu.cn}

\author{Nuowen Kan}
\affiliation{%
  \institution{Shanghai Jiao Tong University}
  \city{Shanghai}
  \country{China}}
\email{kannw_1230@sjtu.edu.cn}

\author{Chenglin Li}
\affiliation{%
  \institution{Shanghai Jiao Tong University}
  \city{Shanghai}
  \country{China}}
\email{lcl1985@sjtu.edu.cn}

\author{Hongkai Xiong}
\affiliation{%
  \institution{Shanghai Jiao Tong University}
  \city{Shanghai}
  \country{China}}
\email{xionghongkai@sjtu.edu.cn}

\thanks{$^\ast$Both authors contributed equally to this research. $^\dagger$Corresponding authors.}

\renewcommand{\shortauthors}{Junyu Zhou et al.}

\begin{abstract}
  Recent prominence in 3D Gaussian Splatting (3DGS) has enabled real-time rendering while maintaining high-fidelity novel view synthesis. However, 3DGS resorts to the Gaussian function that is low-pass by nature and is restricted in representing high-frequency details in 3D scenes. Moreover, it causes redundant primitives with degraded training and rendering efficiency and excessive memory overhead. To overcome these limitations, we propose 3D Gabor Splatting (3DGabSplat) that leverages a novel 3D Gabor-based primitive with multiple directional 3D frequency responses for radiance field representation supervised by multi-view images. The proposed 3D Gabor-based primitive forms a filter bank incorporating multiple 3D Gabor kernels at different frequencies to enhance flexibility and efficiency in capturing fine 3D details. Furthermore, to achieve novel view rendering, an efficient CUDA-based rasterizer is developed to project the multiple directional 3D frequency components characterized by 3D Gabor-based primitives onto the 2D image plane, and a frequency-adaptive mechanism is presented for adaptive joint optimization of primitives. 3DGabSplat is scalable to be a plug-and-play kernel for seamless integration into existing 3DGS paradigms to enhance both efficiency and quality of novel view synthesis. Extensive experiments demonstrate that 3DGabSplat outperforms 3DGS and its variants using alternative primitives, and achieves state-of-the-art rendering quality across both real-world and synthetic scenes. Remarkably, we achieve up to 1.35 dB PSNR gain over 3DGS with simultaneously reduced number of primitives and memory consumption. 
\end{abstract}



\maketitle

\section{Introduction}
Novel view synthesis for complex 3D scenes remains a fundamental challenge in 3D computer vision and graphics. It is indispensable for broad applications ranging from autonomous driving, robot navigation to virtual reality. Traditional methods leverage point clouds and meshes \cite{schonberger2016sfm, goesele2007mvs} to enable explicit scene representation for fast rendering but often lack visual fidelity.
With the rise of deep learning, Neural Radiance Fields (NeRF) \cite{mildenhall2021nerf} pioneers in integrating implicit scene representation with differentiable volumetric rendering to model both geometric structures and view-dependent features via neural networks and allows high-fidelity novel view synthesis. However, NeRF and its variants \cite{barron2021mipnerf, barron2022mipnerf360, barron2023zipnerf} are usually restricted in real-time rendering due to low training efficiency.

\begin{figure}[!t]
\renewcommand{\baselinestretch}{1.0}
\setlength{\abovecaptionskip}{0pt}
\centering
\includegraphics[width=0.9\columnwidth]{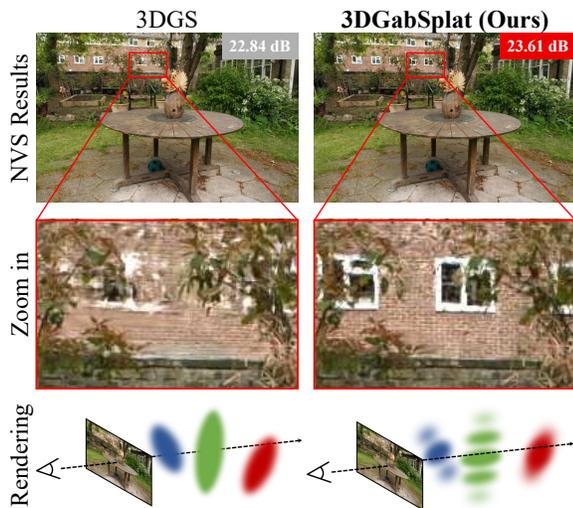}
\caption{Comparison of 3D Gabor Splatting and Gaussian Splatting \cite{kerbl20233dgs}. We propose 3D Gabor-based primitives as a superior alternative to Gaussian kernels, incorporating directional frequencies to more effectively capture high-frequency details in 3D scenes, thereby enabling efficient and high-fidelity novel view synthesis.}\label{fig:intro}
\end{figure}

Recently, 3D Gaussian Splatting (3DGS) \cite{kerbl20233dgs} has gained prominence as an efficient alternative to NeRF that utilizes 3D Gaussian ellipsoids defined on point clouds for explicit scene representation. It simultaneously enables real-time rendering and preserves high-quality performance by projecting the 3D Gaussians onto the 2D image plane and introducing a tile-based rasterizer for parallelized processing.
Inspired by its remarkable success, 3DGS has been extended to a wide range of downstream tasks, including SLAM \cite{matsuki2024gsslam, yan2024gsslam, keetha2024splatam}, autonomous driving \cite{zhou2024drivinggaussian, zhou2024hugs, yan2024street}, large-scale city reconstruction \cite{xiangli2022bungeenerf, lu2023ommo}, dynamic scene reconstruction \cite{yang2024deformable, huang2024scgs, luiten2024dynamic, yang2023gs4d}, surface reconstruction \cite{guedon2024sugar, huang20242dgs, yu2024gof, yu2024gsdf}, and 3D generation \cite{chen2024text23d, tang2023dreamgaussian, yi2023gaussiandreamer}.
However, due to the inherently low-frequency nature of Gaussian functions, 3DGS struggles to represent and reconstruct high-frequency regions with intricate details and complex patterns. It remains unsolved to develop an alternative to Gaussian-based primitives for exploiting appropriate frequency information and representing high-frequency details.

From the signal processing perspective, the Gaussian function serves as a low-pass filter, with its Fourier transform also exhibiting exponential decay along the frequency axis.
In contrast, the Gabor function defined by a Gaussian envelope modulated by a complex exponential term allows band-pass filtering for efficient representation of signals across multiple frequencies and orientations. Gabor filters have been widely integrated into convolutional neural networks (CNNs) for 2D image representation \cite{luan2018gaborCNN, liu2020naive, perez2020gabor, zhu2023learning} and radiance field modeling \cite{fathony2020multiplicative, saragadam2023wire, chen2023neurbf}. Recently, 2D Gabor splatting (2DGabSplat) \cite{wurster20242dgabor} considers Gabor kernels to modulate the Gaussian envelope along fixed direction for high-quality image representation but neglects the directionality in 3D scenes. It is limited by 1D frequency modulation with fixed direction and cannot fully exploit 3D frequency components and optimize the 3D-to-2D projection for 3D scene representation.

To address these issues, we propose \textit{3D Gabor Splatting (3DGabSplat)}, which leverages 3D Gabor-based primitives to achieve frequency adaptive representation for 3D radiance fields.  
The proposed 3D Gabor-based primitive forms a filter bank that collects a Gaussian kernel representing the low-frequency component and multiple 3D Gabor kernels for representing details of varying directions and frequencies. Furthermore, for rendering novel views, directional 3D frequency components characterized by the 3D Gabor-based primitive are projected to 2D image spaces with a specifically designed CUDA-based rasterizer and jointly optimized with a frequency-adaptive framework. 
To the best of our knowledge, 3DGabSplat is the first to incorporate a series of directional 3D frequency components with 3D Gabor kernel based primitives for novel view synthesis. The contributions of this paper are summarized below.
\begin{itemize}
\item We propose 3D Gabor Splatting (3DGabSplat), the first approach that leverages 3D Gabor-based primitives to constitute an adaptive filter bank of multiple directional 3D frequency responses to represent high-frequency details in 3D radiance fields.
\item We develop a differentiable CUDA-based rasterizer for integrating multiple directional frequency responses to project 3D Gabor-based primitives onto the 2D image plane, and present a frequency-adaptive mechanism to jointly optimize the frequency distribution of the primitives.
\item 3DGabSplat achieves state-of-the-art performance in novel view synthesis across both real-world and synthetic scenes, and outperforms 3D Gaussian Splatting with fewer primitives and lower memory consumption.
\end{itemize}

To be concrete, distinguishing from 3DGS and 2D Gabor splatting with 1D frequency modulation, we parameterize each 3D Gabor kernel by distinct 3D frequencies and corresponding weighted coefficients to form the primitive for capturing high-frequency and directional details in 3D scenes. Remarkably, the Gaussian kernel and Gabor kernel with 1D frequency modulation are special cases degenerating in the direction and frequency.
Furthermore, we make specific designs of efficient CUDA-based rasterizer and frequency-adaptive optimization for rendering novel views with the proposed 3D Gabor-based primitives.
The Gabor kernel with its frequency vector is first projected from world coordinates to ray space and then integrated along the z-axis for 3D-to-2D splatting, followed by front-to-back alpha blending. The frequency-adaptive strategy then allows the proposed primitives to dynamically adjust their frequencies and coefficients during densification and optimization to enhance the efficiency of 3D scene representation.

Extensive experiments on real-world and synthetic scenes demonstrate that 3DGabSplat outperforms 3DGS and its extensions, and achieves state-of-the-art performance in novel view synthesis. Moreover, it is evidently superior in capturing complex textures and details, as depicted in \figurename~\ref{fig:intro}. Furthermore, 
3DGabSplat outperforms existing alternatives to Gaussian kernels, and allows real-time rendering and low memory consumption. 
 
\begin{figure*}[!t]
\renewcommand{\baselinestretch}{1.0}
\setlength{\abovecaptionskip}{0pt}
\centering
\begin{subfigure}{\linewidth}
\centering
\includegraphics[width=0.9\linewidth]{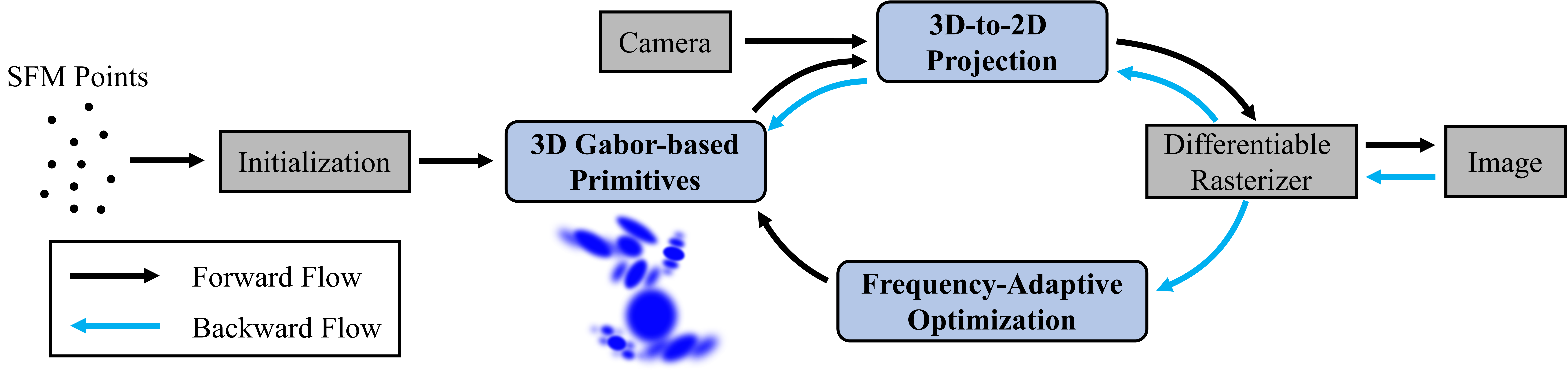}
\caption{The overall training framework of 3DGabSplat.}
\end{subfigure}
\begin{subfigure}{0.64\linewidth}
\centering
\includegraphics[width=\linewidth]{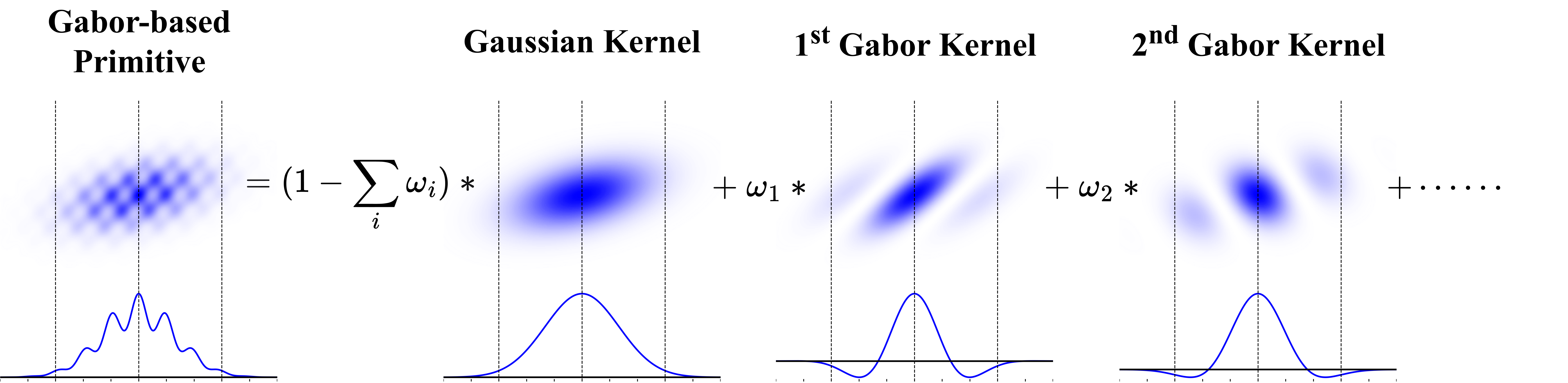}
\caption{Illustration of Gabor-based primitive.}
\end{subfigure}
\hfill
\begin{subfigure}{0.3\linewidth}
\centering
\includegraphics[width=\linewidth]{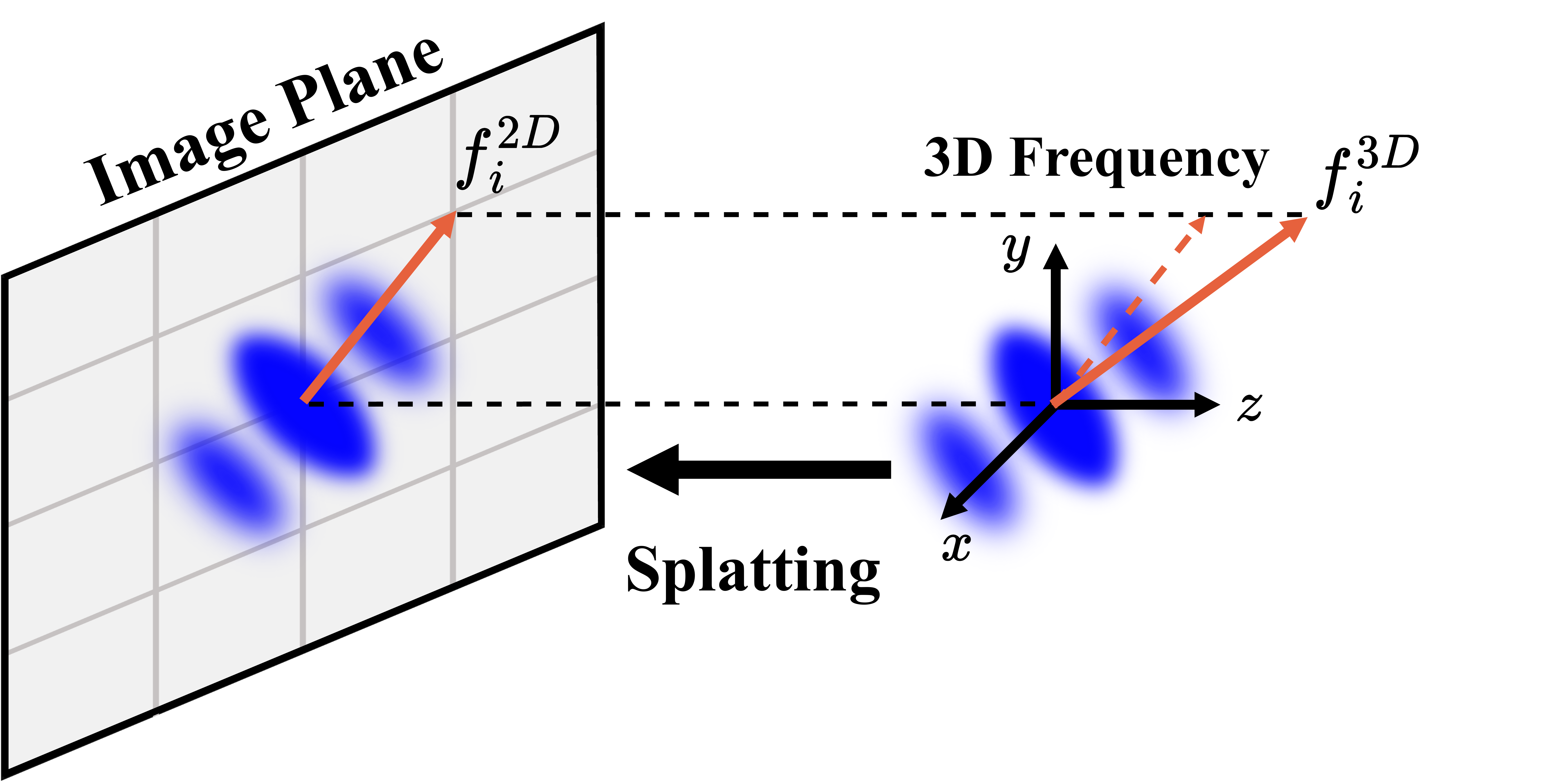}
\caption{Projection from 3D to 2D image plane.}
\end{subfigure}
\caption{Overview of the proposed 3DGabSplat. (a): Training pipeline of 3DGabSplat, with core processes highlighted in blue, including the construction of the 3D Gabor-based primitive, the 3D-to-2D projection of the primitive, and frequency-adaptive optimization strategy. (b): Proposed Gabor-based primitive, formulated as a weighted sum of a Gaussian kernel and Gabor kernels at different frequencies. (c): Projection of the 3D Gabor kernel onto the 2D image plane, where the corresponding 3D frequencies are mapped to 2D vectors.}\label{fig:method}
\end{figure*}

\section{Related Work}
\subsection{Novel View Synthesis}
Novel view synthesis (NVS) aims to generate unseen views of a scene or object through a set of multi-view images.
Neural Radiance Fields (NeRF) \cite{mildenhall2021nerf} revolutionize the field by employing multi-layer perceptrons (MLPs) to model geometry and view-dependent features, which are optimized through volumetric rendering to generate photorealistic images.
The success of NeRF inspires numerous follow-up studies on anti-aliasing \cite{barron2021mipnerf, barron2022mipnerf360, barron2023zipnerf, hu2023trimipnerf}, few-shot reconstruction \cite{jain2021putdiet, niemeyer2022regnerf, yang2023freenerf}, and dynamic scenes \cite{park2021nerfies, fridovich2023kplane, cao2023hexplane}.
However, NeRF and its extensions are limited in real-time applications, due to extensive training and rendering time. Grid-based representations are employed in \cite{fridovich2022plenoxels, muller2022ingp, chen2022tensorf, sun2022dvgo} to accelerate training but often compromise rendering fidelity. Recently, 3D Gaussian Splatting (3DGS) \cite{kerbl20233dgs} has demonstrated high-quality and real-time rendering performance. 
In this paper, we adopt the explicit scene representation of 3DGS rather than neural networks like NeRF and its variants, significantly improving training efficiency and rendering speed.

\subsection{Differentiable Point-based Rendering}
Differentiable point-based rendering \cite{gross2011point, yifan2019differentiable, aliev2020neural, kopanas2021point, lassner2021pulsar} has been widely studied due to its efficiency and flexibility in representing geometry and appearance from image supervision.
The seminal work of 3DGS \cite{kerbl20233dgs} replaces traditional point primitives with 3D Gaussian ellipsoids, the properties of which can be optimized via a multi-view photometric loss. 3DGS offers an explicit scene representation, where each Gaussian can be efficiently splatted onto screen space, thereby enabling real-time rendering performance.
Extensive efforts have been made to enhance the rendering quality of 3DGS, including anti-aliasing \cite{yu2024mipsplat, yan2024msgs, liang2024anasplat}, optimized densification processes \cite{ye2024absgs, yu2024gof, zhang2024fregs, rota2024revisedensify, zhang2024pixelgs, kheradmand2024mcmc}, and grid-based representations \cite{lu2024scaffoldgs, ren2024octreegs}.
Some studies have also investigated the compression and pruning of 3DGS primitives to enhance rendering efficiency \cite{lee2024compactgs, fang2024minisplat, niemeyer2024radsplat, fan2024lightgaussian}.

However, due to the intrinsic low-frequency characteristics of the Gaussian kernel, 3DGS is constrained in its ability to represent and render regions with high-frequency details.
Redundant Gaussian primitives are employed to represent the 3D scene, which inevitably affects computational efficiency and memory overhead.
Alternative primitives have been explored to enhance geometric representation, including generalized exponential functions \cite{hamdi2024ges}, 2D surfels \cite{huang20242dgs}, 3D smooth convexes \cite{held20243dcs}, 3D Half-Gaussian kernels \cite{li20253dhgs}, deformable Beta Kernels \cite{liu2025dbs}.
However, some methods retain the Gaussian kernel to conform with the 3DGS rasterization pipeline, while others produce fragmented reconstructions due to their use of polyhedral representations, which are inherently limited in capturing intricate high-frequency details in 3D scenes and often deliver inferior rendering performance compared to 3DGS.

In this paper, we introduce 3D Gabor-based primitive as a novel kernel for real-time NVS. Multiple directional frequency components are incorporated to construct a filter bank for each primitive, demonstrating a superior alternative to Gaussian-based kernels and achieving both efficient and photorealistic rendering results.

\section{Method}
In this section, we first provide an overview of 3DGS and highlight the motivation of our method. We then introduce the proposed 3D Gabor Splatting and elaborate key components, including construction of 3D Gabor-based primitives and its relations to existing works, projection from 3D world coordinates to the 2D image plane, and frequency-adaptive optimization, as illustrated in \figurename~\ref{fig:method}.

\subsection{Preliminaries and Motivation}
3DGS \cite{kerbl20233dgs} represents the 3D scene with a set of 3D Gaussian primitives $\{ \mathcal{G}_k|k=1,\dots,N\}$, where the $k$-th primitive $\mathcal{G}_k$ is parameterized by the center position (mean) $\boldsymbol{\mu}_k\in \mathbb{R}^3$, 3D covariance matrix $\mathbf{\Sigma}_k\in\mathbb{R}^{3\times 3}$, opacity $\alpha_k$, and spherical harmonics coefficients for color $\mathbf{c}_k$. Mathematically, $\mathcal{G}_k$ is formulated as
\begin{equation}\label{equ:3d_gaussian}
\mathcal{G}_k(\mathbf{x}) = \exp{\left(-\frac{1}{2}(\mathbf{x} - \boldsymbol{\mu}_k)^{\top} \mathbf{\Sigma}_k^{-1} (\mathbf{x} - \boldsymbol{\mu}_k)\right)}.
\end{equation}
In \eqref{equ:3d_gaussian}, the covariance matrix $\mathbf{\Sigma}_k$ is positive semi-definite and can be parameterized as $\mathbf{\Sigma}_k = \mathbf{R}_k \mathbf{S}_k \mathbf{S}_k^\top\mathbf{R}_k^\top$ with the scaling matrix $\mathbf{S}_k$ and rotation matrix $\mathbf{R}_k$.

To render an image, the 3D Gaussian primitives are projected from the world coordinate system onto the 2D image plane to produce corresponding 2D Gaussian primitives.
For $\mathcal{G}_k$, its covariance matrix $\Sigma_k$ is transformed into the camera coordinate system using a world-to-camera transformation $\mathbf{W}$ and transformed to the ray space via a local affine transformation defined by the Jacobian $\mathbf{J}$.
\begin{equation}
\mathbf\Sigma^{\prime}_k=\mathbf{JW}\mathbf\Sigma_k\mathbf{W}^{\top}\mathbf{J}^{\top}.
\end{equation}
Subsequently, the transformed 3D Gaussian primitive is integrated along the $z$-axis, where the top-left 2$\times$2 submatrix of $\mathbf{\Sigma}^{\prime}_k$ is extracted as the 2D covariance matrix $\mathbf{\Sigma}_k^{2D}$.
The 2D Gaussian primitive $\mathcal{G}_k^{2D}$ is formulated using $\mathbf{\Sigma}_k^{2D}$ and the projected center $\boldsymbol{\mu}_k^{2D}$as 
\begin{equation}
\mathcal{G}_k^{2D}(\mathbf{x}) = \exp{\left(-\frac{1}{2} (\mathbf{x} - \boldsymbol{\mu}_k^{2D})^\top (\mathbf{\Sigma}_k^{2D})^{-1} (\mathbf{x} - \boldsymbol{\mu}_k^{2D})\right)}.
\label{equ:2d_gaussian}
\end{equation}
Finally, the splatted 2D Gaussians are sorted by depth, and the color of each pixel is integrated using front-to-back $\alpha$-blending.
\begin{equation}
\mathbf{C}(\mathbf{x})=\sum_{i=1}^N \mathbf{c}_i\alpha_i\mathcal{G}_i^{2D}(\mathbf{x})\prod_{j=1}^{i-1}(1-\alpha_j\mathcal{G}_j^{2D}(\mathbf{x})).
\label{equ:alpha-blend-gauss}
\end{equation}

However, each primitive in 3DGS is modeled with a single Gaussian kernel, whose inherent low-pass nature limits its ability to represent high-frequency details. 
This motivates us to incorporate Gabor kernels as band-pass components for each primitive. A Gabor function is defined as the product of a Gaussian function and a complex sinusoidal wave. Its general form in one dimension (1D) is 
\begin{equation}
g(t)=\frac{1}{\sqrt{2\pi}\sigma}\exp{(-\frac{t^2}{2\sigma^2})}\cdot \exp{(j2\pi ft)},
\label{equ:1d_gabor}
\end{equation}
where $f$ denotes the center frequency of the Gabor filter, and $\sigma$ represents the variance of the Gaussian function. 
In the subsequent section, we extend the 1D Gabor kernel to 3D space and employ it to construct 3D primitives for efficient representation of high-frequency structures and directional information in 3D scenes.

\begin{figure}[!t]
\renewcommand{\baselinestretch}{1.0}
\setlength{\abovecaptionskip}{0pt}
\centering
\includegraphics[width=\columnwidth]{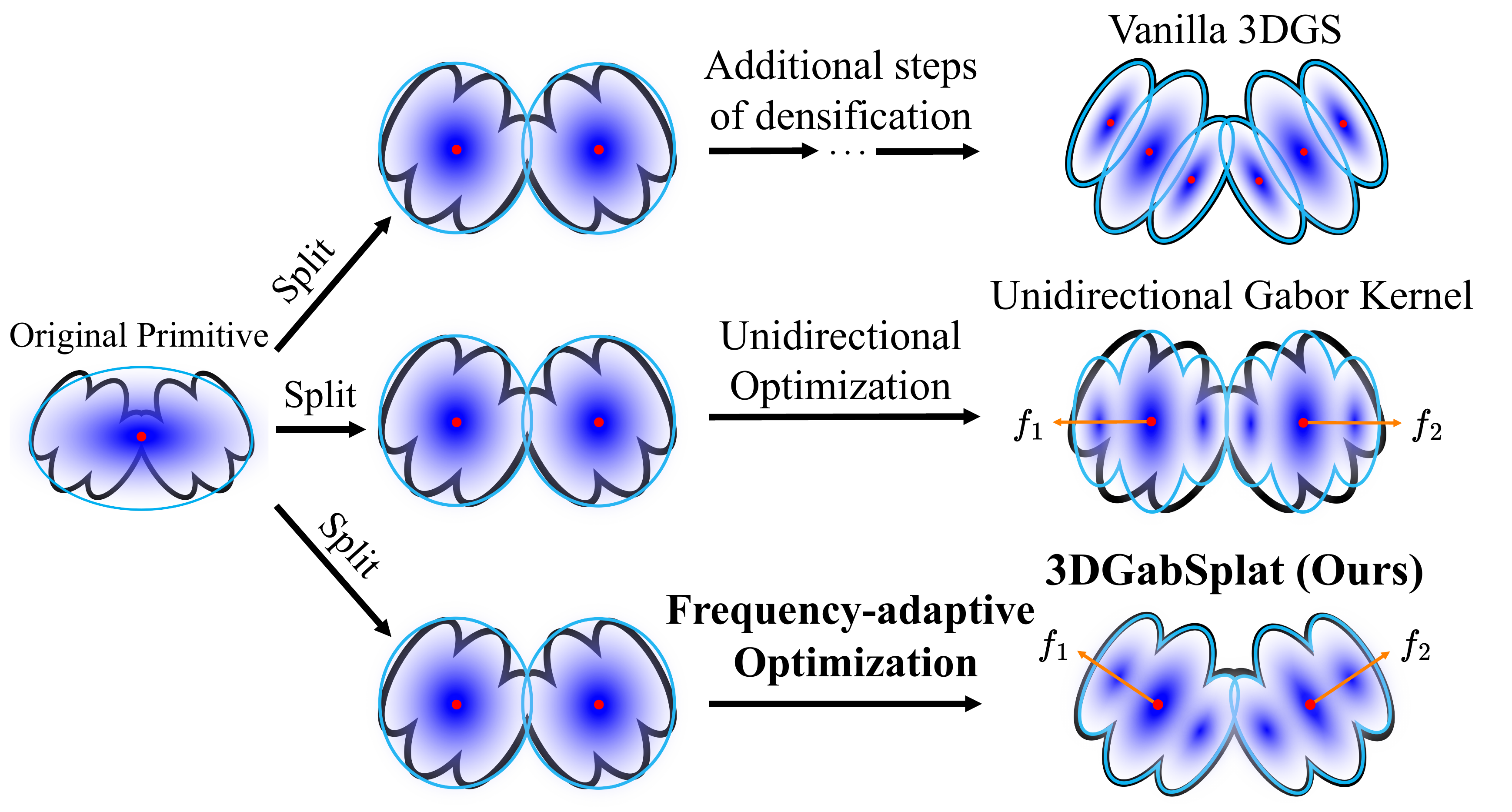}
\caption{Comparison of our 3DGabSplat with 3DGS and 2DGabSplat. 
Beginning with a single Gaussian primitive, vanilla 3DGS requires iterative densification, ultimately expanding to six Gaussian kernels. 
When using unidirectional Gabor kernel, only two primitives are required, but they fail to fully capture the complex pattern. 
In contrast, our method employs a single splitting step followed by frequency-adaptive optimization, allowing our Gabor-based primitives to effectively represent complex patterns with only two primitives.
}
\label{fig:kernel_comparison}
\end{figure}

\subsection{3D Gabor-based Primitives} %
In this section, we first provide a detailed description of our proposed 3D Gabor-based primitives for novel view synthesis. Subsequently, we clarify the relation between our 3D Gabor-based primitives and existing methods.

\subsubsection{Construction of 3D Primitives}
We extend the conventional 1D Gabor function in \eqref{equ:1d_gabor} to its 3D form. To ensure computational efficiency while preserving frequency information, we retain only the real part of the Gabor function, and yield the cosine term as
\begin{equation}\label{equ:3d_gabor}
g(\mathbf{x})=\frac{1}{(2\pi)^{3/2}|\mathbf{\Sigma}|^{1/2}}\exp{\left(-\frac{1}{2}\mathbf{x}^T\mathbf{\Sigma}^{-1}\mathbf{x}\right)}\cdot\cos(2\pi\boldsymbol{f}^\top \mathbf{x}),
\end{equation}
where $\boldsymbol{f}\in\mathbb{R}^3$ is the vector of center frequencies along the three dimensions. 

Since Gaussian-based primitives are inherently low-pass filters that are limited in representing and rendering high-frequency details, we incorporate high-frequency information to enhance their ability to capture and represent 3D scenes.
To this end, we integrate Gabor filters with varying frequencies as band-pass components for each primitive, alongside the Gaussian kernel, to form a filter bank.
Building upon \eqref{equ:3d_gaussian}, the proposed 3D Gabor-based primitive is formulated as a weighted sum of a Gaussian kernel and Gabor kernels with varying center frequencies.
\begin{equation}
\{Gabor\}_k(\mathbf{x})=\mathcal{G}_k(\mathbf{x})\cdot \Big[\big(1-\sum_{i=1}^F \omega_{k,i}\big) + \sum_{i=1}^F \omega_{k,i}\cos\big(2\pi \boldsymbol{f}_{k,i}^\top(\mathbf{x}-\boldsymbol{\mu}_k)\big)\Big].
\end{equation}
Here, $\boldsymbol{f}_{k,i}|i=1,..,F$ denotes the center frequencies of the $F$ distinct Gabor kernels for primitive $\{Gabor\}_k$, while $\omega_{k,i}$ represents the corresponding weight coefficients, ensuring that the sum of the weight coefficients for all Gabor kernels and the Gaussian kernel equals 1.
Similar to other attributes of the primitive, $\boldsymbol{f}_{k,i}$ and $\omega_{k,i}$ are optimized and updated during training through multi-view photometric supervision.
Each primitive is assigned distinct frequencies to meet the representation and rendering requirements of its specific region.

\subsubsection{Relation to Existing Works} Gaussian-based primitives in 3DGS can be viewed as a special case of the proposed 3D Gabor-based primitives with $\boldsymbol{f}_{k,i} = 0$ and $\omega_{k,i} = 0$. Moreover, 2DGabSplat \cite{wurster20242dgabor} utilizes a unidirectional Gabor kernel with fixed frequency intervals to construct 2D primitives, as formulated below.
\begin{equation}
g(\mathbf{x})=\exp\left(-\frac{1}{2}\mathbf{x}^T\mathbf{\Sigma}^{-1}\mathbf{x}\right)\cos\left(2\pi f_dx_d\right),
\end{equation}
where $f_d$ denotes the unidirectional frequency along $x_d$. It is also a degenerate case of the proposed 3D Gabor-based primitive constrained by 1D frequency modulation with a fixed orientation. In \figurename~\ref{fig:kernel_comparison}, we present a visual comparison to demonstrate that, compared with the Gaussian-based primitive and the unidirectional 2D Gabor-based primitive, the proposed 3D Gabor-based primitives achieve efficient and precise representation of complex patterns with fewer primitives.

\subsection{3D Gabor Splatting}
Building upon the proposed 3D Gabor-based primitive, we develop 3D Gabor Splatting for real-time novel view synthesis. The overall pipeline consists of transformation from world coordinates to ray space, integrating along the z-axis to splat onto the 2D image plane, and performing tile-based color-alpha blending for rasterization.

\subsubsection{Coordinate Transformation of 3D Primitives}
Firstly, since the 3D Gabor-based primitive can be represented as a weighted sum of Gabor filters at different frequencies, it follows that each individual Gabor filter can be analyzed and projected onto the image plane separately.
Through derivation, it is demonstrated that the projection of the Gabor primitive is equivalent to the projection of its 3D frequency, employing a strategy analogous to the projection of the 3D covariance matrix. The detailed procedure is outlined in the supplementary material. By utilizing the world-to-camera transformation and the local affine Jacobian matrix, the projected frequency can be expressed as
\begin{equation}
\boldsymbol{f}_{k,i}^{proj} = {(\mathbf{JW})^{-1}}^\top  \boldsymbol{f}_{k,i}.
\end{equation}

\subsubsection{3D-to-2D Splatting}
After projection to the ray space, the 3D Gabor-based primitive is integrated along the z-axis to achieve the 3D-to-2D splatting. 
The integration results in the Gaussian covariance matrix retaining only its upper-left $2\times 2$ submatrix, while the splatted 2D frequency ${\boldsymbol{f}_{k,i}^{2D}}$ is influenced by the frequency component along the $z$-axis, which can be expressed as
\begin{equation}\left\{\begin{aligned}
{{f}_{k,i,x}^{2D}} &= {f}_{k,i,x}^{proj} - \frac{S_{02}}{S_{22}}\cdot{f}_{k,i,z}^{proj}\\
{{f}_{k,i,y}^{2D}} &= {f}_{k,i,y}^{proj} - \frac{S_{12}}{S_{22}}\cdot{f}_{k,i,z}^{proj}
\end{aligned}\right.,
\end{equation}
where $S_{02}, S_{12}, S_{22}$ are elements from the inverse of the Gaussian covariance matrix
\begin{equation}
\boldsymbol{\Sigma}_k^{-1}=\begin{pmatrix}S_{00}&S_{01}&S_{02}\\S_{10}&S_{11}&S_{12}\\S_{20}&S_{21}&S_{22}\end{pmatrix}.
\end{equation}
The detailed derivation is also presented in the supplementary material. After splatting, the 2D Gabor-based primitive can be formulated as the product of a 2D Gaussian kernel $\mathcal{G}_k^{2D}(\mathbf{x})$ and the weighted frequency coefficients.
\begin{equation}
\{Gabor\}_k^{2D\!}(\mathbf{x})\!=\!\mathcal{G}_k^{2D\!}(\mathbf{x})\!\left[\!(1\!-\!\sum_{i=1}^F\!\omega_{k,i})\!+\!\sum_{i=1}^F\!\omega_{k,i}\!\cos(2\pi {\boldsymbol{f}_{k,i}^{2D}}^{\top\!}\!(\mathbf{x}\!-\!\boldsymbol{\mu}_k^{2D\!})\!)\!\right]\!.
\end{equation}

\subsubsection{Rasterizer of 3DGabSplat}
To enable real-time rendering, we adopt the tile-based rasterizer from 3DGS and adapt it to the proposed primitives. Compared to \eqref{equ:alpha-blend-gauss}, we replace the gaussian kernel with the 2D Gabor-based primitives ${Gabor}_i^{2D}(\mathbf{x})$. The resulting color $\alpha$-blending for each pixel in the image is then formulated as:
\begin{equation}
\mathbf{C}(\mathbf{x})=\sum_{i=1}^N\mathbf{c}_i\alpha_i\{Gabor\}_i^{2D}(\mathbf{x})\prod_{j=1}^{i-1}(1-\alpha_j\{Gabor\}_j^{2D}(\mathbf{x})).
\end{equation}

\subsection{Optimization}
\subsubsection{Initialization and Loss}
The optimization process starts by generating a sparse point cloud using Structure-from-Motion (SFM) \cite{schonberger2016sfm}.
Each point is assigned with Gaussian attributes as in 3DGS, including the center position $\boldsymbol{\mu}_k$, covariance matrix $\mathbf{\Sigma}_k$, opacity $\alpha_k$, and SH coefficients $\mathbf{c}_k$.
Additional parameters, including 3D frequencies $\boldsymbol{f}_{i}|\ i=1,\dots,F$ and corresponding weighting coefficients $\omega_{k,i}$, are incorporated to formulate the 3D Gabor-based primitive.
We initialize the frequencies and weighting coefficients of all Gabor kernels to small values, such as 0.001 and 0.01, respectively, ensuring a progressive training process from low to high frequencies.
Notably, the frequencies and coefficients are deliberately nonzero at initialization to avoid degeneracy into a Gaussian kernel, which would lead to zero gradients and hinder optimization.
The sigmoid activation function is utilized to constrain the opacity $\alpha_k$ and weighting coefficients $\omega_{k,i}$ within the range $[0,1)$, thereby ensuring stable training.
During training, we adopt the loss function as in 3DGS which integrates $\mathcal{L}_1$ with a D-SSIM term
\begin{equation}
\mathcal{L}=(1-\lambda)\mathcal{L}_1+\lambda\mathcal{L}_\mathrm{D-SSIM},
\end{equation}
where we set $\lambda = 0.2$ for all experiments.

\subsubsection{Frequency-adaptive Optimization}
The initial sparse point cloud generated by SFM is inadequate for accurately representing the 3D scene. 3DGS incorporates an adaptive density control mechanism to optimize Gaussian primitives through densification and pruning.
Additional Gaussian primitives are introduced through splitting or cloning during the densification process when the cumulative average view-space positional gradient exceeds a predefined threshold.
However, adopting the densification approach as in 3DGS results in the newly generated child primitives being simple clones of the original primitives. This is detrimental to our 3D Gabor-based primitives, as the frequencies and coefficients escalate with densification, causing an overaccumulation of high-frequency primitives that degrade the rendering quality.

\begin{table*}[!t]
\renewcommand{\baselinestretch}{1.0}
\renewcommand{\arraystretch}{1.0}
\setlength{\abovecaptionskip}{0pt}
\centering
\caption{Quantitative results on Mip-NeRF360 \cite{barron2022mipnerf360}, Tanks \& Temples \cite{knapitsch2017tnt}, and Deep Blending \cite{hedman2018db}. The 1st, 2nd, and 3rd best performances in each column are highlighted in red, orange, and yellow, respectively. Note that * indicates reproduced results.}
\label{tab:main_res_table}
\begin{tabular}{@{}l|ccc|ccc|ccc@{}}
\toprule
Dataset & \multicolumn{3}{c|}{Mip-NeRF360} & \multicolumn{3}{c|}{Tanks\&Temples} & \multicolumn{3}{c}{Deep Blending} \\
Method | Metrics & SSIM$\uparrow$ & PSNR (dB)$\uparrow$ & LPIPS$\downarrow$ & SSIM$\uparrow$ & PSNR (dB)$\uparrow$ & LPIPS$\downarrow$ & SSIM$\uparrow$ & PSNR (dB)$\uparrow$ & LPIPS$\downarrow$ \\ \hline
Mip-NeRF360 \cite{barron2022mipnerf360} & 0.792 & 27.69 & 0.237 & 0.759 & 22.22 & 0.257 & 0.901 & 29.40 & 0.245 \\
3DGS \cite{kerbl20233dgs} & 0.815 & 27.21 & 0.214 & 0.841 & 23.14 & 0.183 & 0.903 & 29.41 & 0.243 \\
3DGS* & 0.814 & 27.41 & 0.216 & 0.848 & 23.69 & 0.176 & 0.904 & 29.55 & 0.244 \\
2DGS \cite{huang20242dgs} & 0.804 & 27.03 & 0.239 & 0.832 & 23.16 & 0.212 & 0.903 & 29.50 & 0.257 \\
GES \cite{hamdi2024ges} & 0.794 & 26.91 & 0.250 & 0.836 & 23.35 & 0.198 & 0.901 & 29.68 & 0.252 \\
3DCS \cite{held20243dcs} & 0.802 & 27.29 & 0.207 & 0.851 & 23.95 & \cellcolor[HTML]{FFF6B2}0.157 & 0.902 & 29.81 & 0.237 \\
\textbf{3DGabSplat (Proposed)} & 0.818 & \cellcolor[HTML]{FFCC99}27.85 & 0.210 & 0.855 & \cellcolor[HTML]{FFF6B2}24.49 & 0.170 & \cellcolor[HTML]{FFF6B2}0.909 & 30.09 & \cellcolor[HTML]{FFF6B2}0.237 \\
\hline
Scaffold-GS \cite{lu2024scaffoldgs} & 0.811 & 27.72 & 0.228 & 0.853 & 23.96 & 0.177 & 0.906 & 30.21 & 0.254 \\
Scaffold-GS* & 0.815 & 27.72 & 0.220 & 0.852 & 24.00 & 0.175 & \cellcolor[HTML]{FFF6B2}0.909 & \cellcolor[HTML]{FFF6B2}30.32 & 0.253 \\
Octree-GS \cite{ren2024octreegs} & 0.815 & 27.73 & 0.217 & \cellcolor[HTML]{FF9999}0.866 & \cellcolor[HTML]{FF9999}24.52 & \cellcolor[HTML]{FF9999}0.153 & \cellcolor[HTML]{FF9999}0.913 & \cellcolor[HTML]{FFCC99}30.41 & 0.238 \\
Octree-GS* & 0.814 & 27.49 & 0.219 & \cellcolor[HTML]{FFF6B2}0.860 & 24.30 & 0.160 & 0.908 & 30.17 & 0.251 \\
\textbf{3DGabSplat+Scaffold-GS} & 0.815 & \cellcolor[HTML]{FFF6B2}27.84 & 0.220 & 0.859 & \cellcolor[HTML]{FF9999}24.52 & 0.166 & \cellcolor[HTML]{FFCC99}0.911 & \cellcolor[HTML]{FF9999}30.52 & 0.253 \\
\hline
Mip-Splatting \cite{yu2024mipsplat} & 0.817 & 27.58 & 0.218 & 0.851 & 23.79 & 0.178 & 0.905 & 29.69 & 0.248 \\
Analytic-Splatting \cite{liang2024anasplat} & 0.815 & 27.58 & 0.217 & 0.851 & 23.84 & 0.177 & 0.905 & 29.75 & 0.248 \\
AbsGS \cite{ye2024absgs} & 0.820 & 27.49 & 0.191 & 0.853 & 23.73 & 0.162 & 0.902 & 29.67 & \cellcolor[HTML]{FFCC99}0.236 \\
Mip-Splatting+AbsGS* & \cellcolor[HTML]{FFF6B2}0.827 & 27.66 & \cellcolor[HTML]{FFCC99}0.188 & 0.859 & 23.71 & \cellcolor[HTML]{FFF6B2}0.157 & 0.902 & 29.31 & 0.239 \\
Analytic-Splatting+AbsGS* & \cellcolor[HTML]{FFCC99}0.828 & 27.66 & \cellcolor[HTML]{FFF6B2}0.190 & 0.859 & 23.72 & 0.158 & 0.903 & 29.57 & 0.243 \\
\textbf{3DGabSplat+Mip-Splatting+AbsGS} & \cellcolor[HTML]{FF9999}0.829 & \cellcolor[HTML]{FF9999}27.93 & \cellcolor[HTML]{FF9999}0.185 & \cellcolor[HTML]{FFCC99}0.863 & 24.37 & \cellcolor[HTML]{FF9999}0.153 & 0.907 & 29.79 & \cellcolor[HTML]{FF9999}0.231\\
\bottomrule
\end{tabular}
\end{table*}

To address this, we propose a frequency-adaptive optimization scheme for densification. 
Specifically, during densification, we reset the frequency and corresponding weighting coefficients of the newly generated child primitives to small values. The reset mechanism effectively prevents the generation of excessively high-frequency primitives, allowing each newly generated primitive to adaptively adjust its frequency and coefficients during subsequent optimization.
Furthermore, every 3000 epochs, we reset the weighting coefficients of all Gabor kernels in conjunction with opacity reset, effectively eliminating redundant high-frequency primitives. 
The frequency-adaptive optimization strategy enables each primitive to dynamically refine the frequencies and corresponding weighting coefficients of its Gabor kernels, allowing adaptive representation across spatial regions and thereby improving novel view synthesis performance.

\section{Experiments}

\subsection{Experimental Setup}
\textbf{Datasets and Metrics}
To assess the effectiveness of our proposed 3DGabSplat in novel view synthesis, we conduct experiments on both real-world and synthetic datasets, including all scenes from Mip-NeRF 360 (9 scenes) \cite{barron2022mipnerf360}, two from Tanks and Temples (T\&T) \cite{knapitsch2017tnt}, two from Deep Blending (DB) \cite{hedman2018db}, and eight synthetic scenes from the NeRF Synthetic Dataset \cite{mildenhall2021nerf}.
The selected scenes comprise both bounded indoor and unbounded outdoor environments, enabling a comprehensive evaluation of our method's performance.
Consistent with prior work, three widely used metrics in novel view synthesis, PSNR, SSIM \cite{wang2004ssim}, and LPIPS \cite{zhang2018lpips}, are employed to evaluate performance on each dataset.

\noindent \textbf{Baselines}
We begin with a comparative evaluation of our 3DGabSplat against the typical implicit neural rendering method Mip-NeRF360 \cite{barron2022mipnerf360} and state-of-the-art explicit representation methods employing different primitives, including 3DGS \cite{kerbl20233dgs}, 2DGS \cite{huang20242dgs}, GES \cite{hamdi2024ges}, and 3DCS \cite{held20243dcs}.
Furthermore, we conduct experiments based on various extensions of 3DGS, including Scaffold-GS \cite{lu2024scaffoldgs}, Octree-GS \cite{ren2024octreegs}, Mip-Splatting \cite{yu2024mipsplat}, Analytic-Splatting \cite{liang2024anasplat}, and AbsGS \cite{ye2024absgs}. 
To ensure a fair comparison, we integrate implicit structured representations\cite{lu2024scaffoldgs}, anti-aliasing techniques \cite{yu2024mipsplat}, and a modified densification strategy \cite{ye2024absgs} into 3DGabSplat, denoted as 3DGabSplat+Scaffold-GS and 3DGabSplat+Mip-Splatting+AbsGS, by solely replacing Gaussian Splatting with 3DGabSplat.

\begin{figure*}[!t]
\renewcommand{\baselinestretch}{1.0}
\setlength{\abovecaptionskip}{0pt}
\centering
\begin{subfigure}[b]{0.19\textwidth}
\captionsetup{labelformat=empty}
\caption{Ground Truth}
\centering
\includegraphics[width=\textwidth]{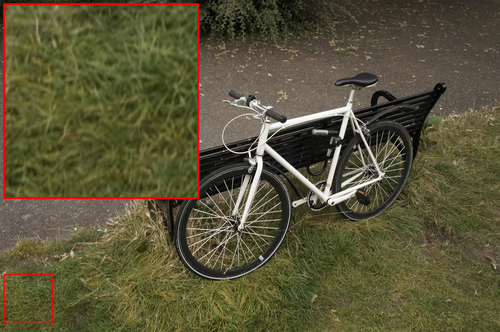}
\end{subfigure}
\begin{subfigure}[b]{0.19\textwidth}
\captionsetup{labelformat=empty}
\caption{3DGabSplat (Ours)}
\centering
\includegraphics[width=\textwidth]{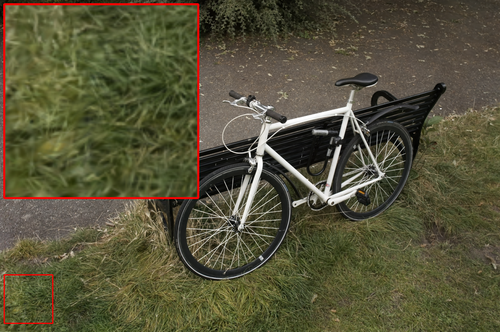}
\end{subfigure}
\begin{subfigure}[b]{0.19\textwidth}
\captionsetup{labelformat=empty}
\caption{3DGS \cite{kerbl20233dgs}}
\centering
\includegraphics[width=\textwidth]{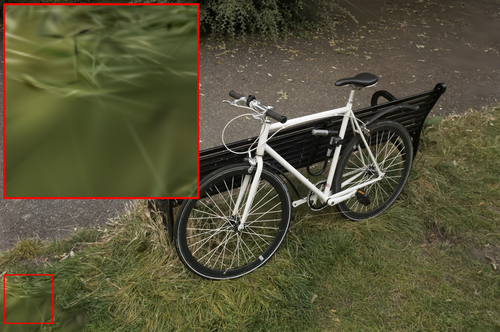}
\end{subfigure}
\begin{subfigure}[b]{0.19\textwidth}
\captionsetup{labelformat=empty}
\caption{GES \cite{hamdi2024ges}}
\centering
\includegraphics[width=\textwidth]{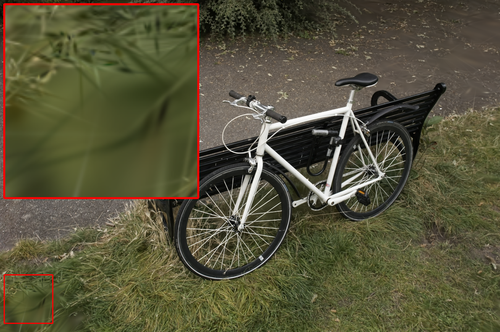}
\end{subfigure}
\begin{subfigure}[b]{0.19\textwidth}
\captionsetup{labelformat=empty}
\caption{3DCS \cite{held20243dcs}}
\centering
\includegraphics[width=\textwidth]{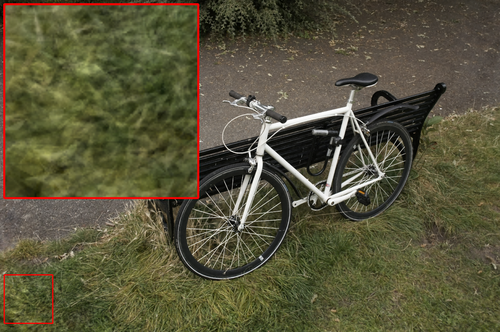}
\end{subfigure}  
\begin{subfigure}[b]{0.19\textwidth}
\centering
\includegraphics[width=\textwidth]{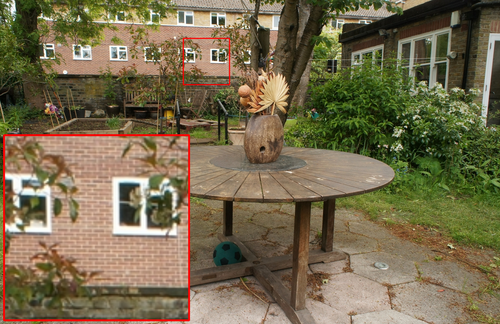}
\end{subfigure}
\begin{subfigure}[b]{0.19\textwidth}
\centering
\includegraphics[width=\textwidth]{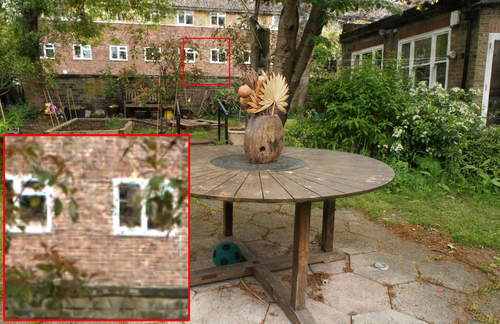}
\end{subfigure}
\begin{subfigure}[b]{0.19\textwidth}
\centering
\includegraphics[width=\textwidth]{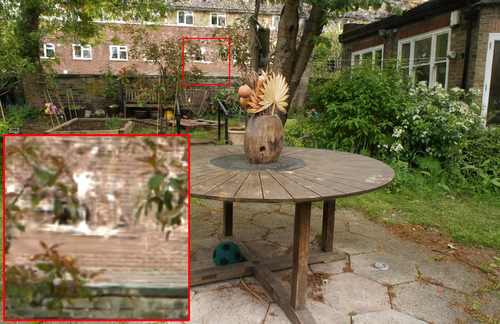}
\end{subfigure}
\begin{subfigure}[b]{0.19\textwidth}
\centering
\includegraphics[width=\textwidth]{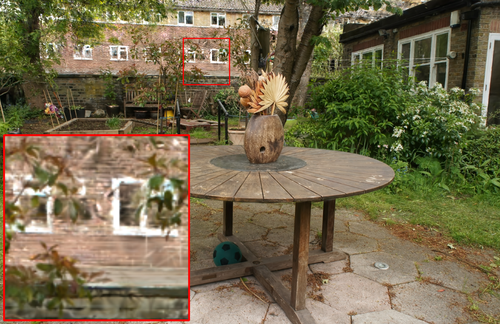}
\end{subfigure}
\begin{subfigure}[b]{0.19\textwidth}
\centering
\includegraphics[width=\textwidth]{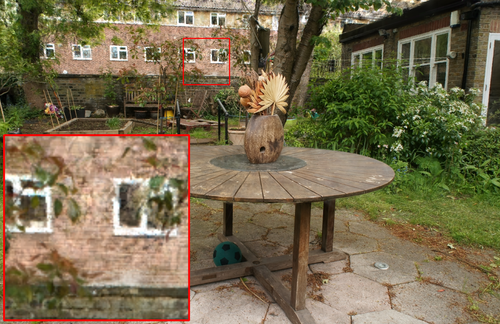}
\end{subfigure}
\begin{subfigure}[b]{0.19\textwidth}
\centering
\includegraphics[width=\textwidth]{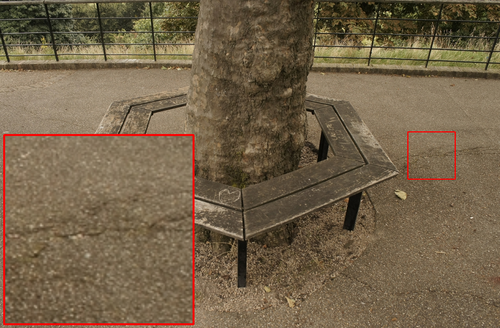}
\end{subfigure}
\begin{subfigure}[b]{0.19\textwidth}
\centering
\includegraphics[width=\textwidth]{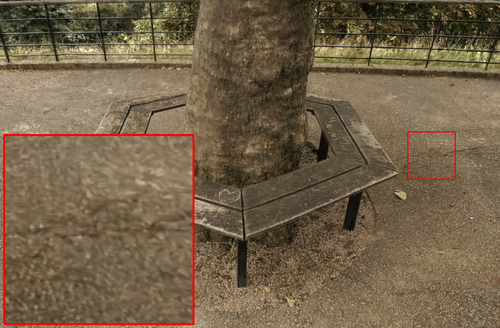}
\end{subfigure}
\begin{subfigure}[b]{0.19\textwidth}
\centering
\includegraphics[width=\textwidth]{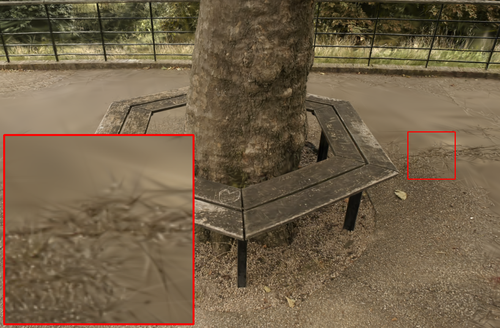}
\end{subfigure}
\begin{subfigure}[b]{0.19\textwidth}
\centering
\includegraphics[width=\textwidth]{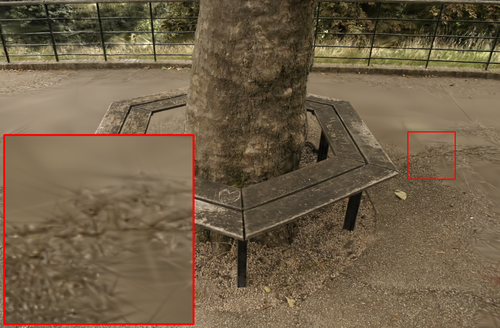}
\end{subfigure}
\begin{subfigure}[b]{0.19\textwidth}
\centering
\includegraphics[width=\textwidth]{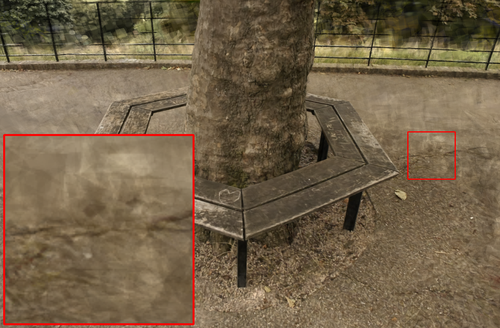}
\end{subfigure}
\begin{subfigure}[b]{0.19\textwidth}
\centering
\includegraphics[width=\textwidth]{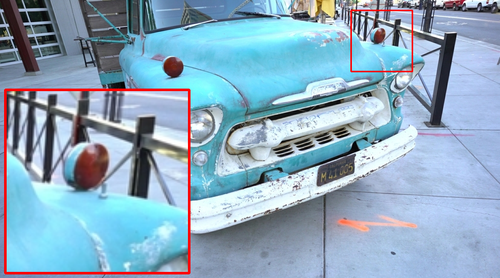}
\end{subfigure}
\begin{subfigure}[b]{0.19\textwidth}
\centering
\includegraphics[width=\textwidth]{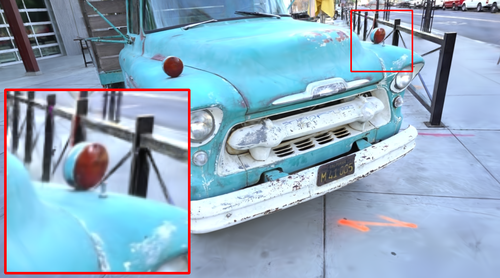}
\end{subfigure}
\begin{subfigure}[b]{0.19\textwidth}
\centering
\includegraphics[width=\textwidth]{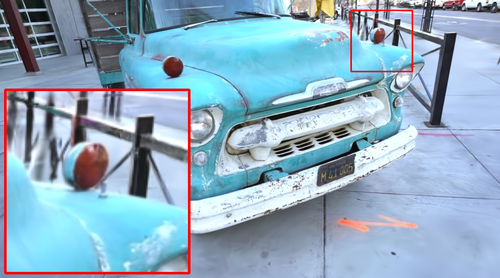}
\end{subfigure}
\begin{subfigure}[b]{0.19\textwidth}
\centering
\includegraphics[width=\textwidth]{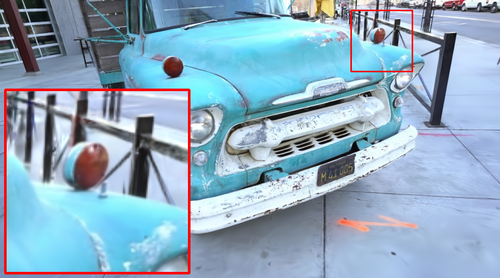}
\end{subfigure}
\begin{subfigure}[b]{0.19\textwidth}
\centering
\includegraphics[width=\textwidth]{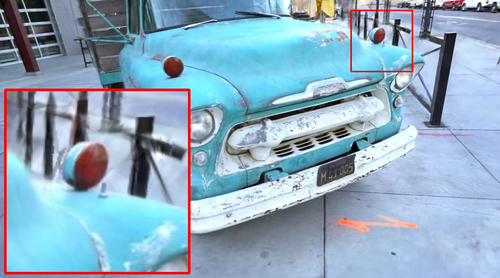}
\end{subfigure}
\begin{subfigure}[b]{0.19\textwidth}
\centering
\includegraphics[width=\textwidth]{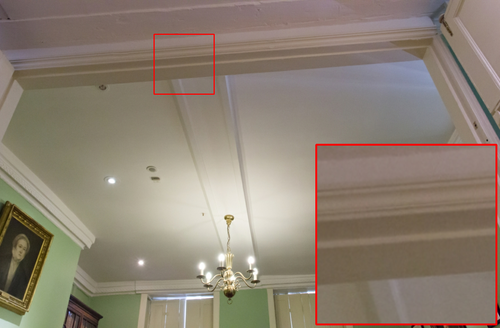}
\end{subfigure}
\begin{subfigure}[b]{0.19\textwidth}
\centering
\includegraphics[width=\textwidth]{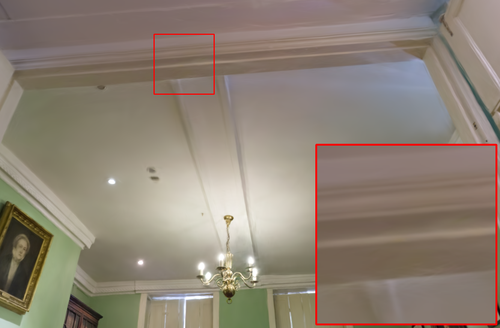}
\end{subfigure}
\begin{subfigure}[b]{0.19\textwidth}
\centering
\includegraphics[width=\textwidth]{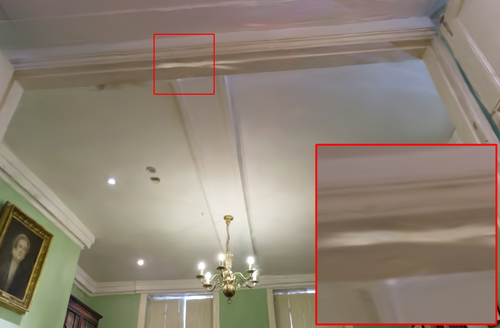}
\end{subfigure}
\begin{subfigure}[b]{0.19\textwidth}
\centering
\includegraphics[width=\textwidth]{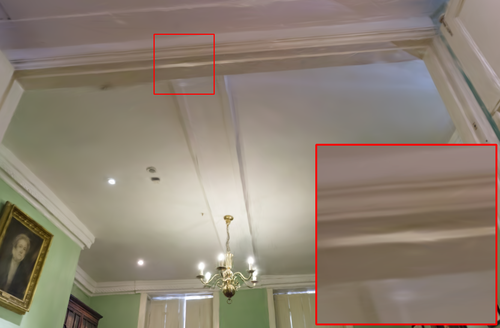}
\end{subfigure}
\begin{subfigure}[b]{0.19\textwidth}
\centering
\includegraphics[width=\textwidth]{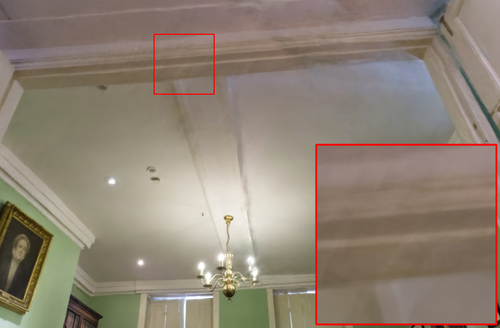}
\end{subfigure}    
\caption{Qualitative evaluation of 3DGabSplat in comparison with 3DGS, GES, and 3DCS. The results demonstrate that our 3DGabSplat outperforms other methods employing different primitives in representing intricate structures, complex textures, and other high-frequency detail regions within 3D scenes.}\label{fig:comparison}
\end{figure*}

\noindent\textbf{Implementation Details}
For the implementation of 3DGabSplat, we set the frequency numbers of the Gabor kernel to $F=2$, where each kernel comprises three-dimensional frequency components along with their associated weighting coefficients, thus an additional $4 \times F$ parameters per primitive are incorporated compared to the Gaussian kernel. In the default configuration, we initialize all frequencies and their corresponding weighting coefficients to 0.001 and 0.01, respectively, thereby ensuring that the primitives exhibit low-frequency characteristics at the onset of training. The learning rates for opacity, frequency, and weighting coefficients are set to 0.025, 0.01, and 0.02, respectively. The other parameters are kept consistent with those of the baseline method to ensure a fair comparison. For the Mip-NeRF 360 dataset, outdoor and indoor scenes are trained at 1/4 and 1/2 resolution  respectively, utilizing the officially provided downsampled versions to ensure fair comparison. All other datasets are trained on images at their original resolution. A more detailed parameter configuration can be found in the supplementary material. All experiments were conducted on an NVIDIA RTX 3090 GPU.

\subsection{Results for Novel View Synthesis}
\subsubsection{Results of Real-World Scenes.}
Quantitative results of real-world scenes are presented in Table \ref{tab:main_res_table}. 
The main results are organized into three parts, corresponding to the top, middle, and bottom blocks in the table.
To begin with, our 3DGabSplat outperforms both the implicit neural rendering method and state-of-the-art 3DGS extensions with different primitives across all three key metrics.
For the PSNR metric, our method achieves improvements of 0.64 dB, 1.35 dB, and 0.68 dB over the 3DGS baseline on the Mip-NeRF 360, T\&T, and DB datasets, respectively.
Meanwhile, our 3DGabSplat, without relying on additional methods, already surpasses the current state-of-the-art 3DGS extension in rendering performance.
As shown in the rendered results in \figurename~\ref{fig:comparison}, our 3DGabSplat exhibits superior performance in capturing intricate structures, complex textures, and high-frequency details compared to methods based on other primitives. This demonstrates the effectiveness of incorporating adaptive frequency information into each primitive. 
In contrast, 3DGS and GES often produce blurry and incomplete reconstructions due to their reliance on low-pass Gaussian kernels, while 3DCS tends to generate fragmented renderings due to its polyhedral representation.
Furthermore, as shown in the middle and bottom part of Table \ref{tab:main_res_table}, both 3DGabSplat+Scaffold-GS and 3DGabSplat+Mip-Splatting+AbsGS exhibit superior performance compared to their respective baselines.
Notably, when integrated with implicit structured representation methods, our 3DGabSplat achieves performance that matches or even exceeds the current state-of-the-art Octree-GS on the T\&T and DB datasets. Additionally, by incorporating anti-aliasing and AbsGrad, our 3DGabSplat establishes a new state-of-the-art on the Mip-NeRF360 dataset, yielding a 0.27 dB improvement in PSNR over Mip-Splatting+AbsGS and a nearly 0.72 dB enhancement compared to the 3DGS baseline.

\subsubsection{Results of Synthetic Scenes}
To further evaluate the 3D reconstruction capability of our 3DGabSplat in bounded scenes, we conduct experiments on the NeRF synthetic dataset, with the quantitative comparison results summarized in Table \ref{tab:syn_res}.
Without incorporating additional methods, our proposed 3DGabSplat achieves superior performance across all metrics, outperforming all competing approaches, including 3DGS variants with different primitives and grid-based representation methods, and achieving a 0.38 dB improvement in PSNR over 3DGS baseline.
We further present a qualitative comparison between our 3DGabSplat and 3DGS in \figurename~\ref{fig:syn_comparison}. 
The results indicate that our method significantly outperforms 3DGS in representing color and texture details as evidenced by the zoom-in region of the mic scene, further underscoring its superior ability to reconstruct high-frequency regions.

\begin{table}[!t]
\renewcommand{\baselinestretch}{1.0}
\renewcommand{\arraystretch}{1.0}
\setlength{\abovecaptionskip}{0pt}
\centering
\caption{Quantitative results on NeRF Synthetic dataset \cite{mildenhall2021nerf}. 
}\label{tab:syn_res}
\begin{tabular}{@{}l ccc@{}}
\toprule
\multicolumn{1}{c}{Method}  & SSIM↑ & PSNR (dB)↑ & LPIPS↓ \\ \hline
Mip-NeRF360 \cite{barron2022mipnerf360} & 0.9610 & 33.090 & 0.0450 \\
3DGS \cite{kerbl20233dgs} & \cellcolor[HTML]{FFCC99}0.9694 & \cellcolor[HTML]{FFCC99}33.347 & \cellcolor[HTML]{FFCC99}0.0302 \\
2DGS \cite{huang20242dgs} & \cellcolor[HTML]{FFF6B2}0.9679 & 33.059 & \cellcolor[HTML]{FFF6B2}0.0330 \\
GES \cite{hamdi2024ges} & 0.9677 & 32.955 & 0.0338 \\
3DCS \cite{held20243dcs} & 0.9558 & 31.134 & 0.0476 \\
Scaffold-GS \cite{lu2024scaffoldgs} & 0.9674 & \cellcolor[HTML]{FFF6B2}33.135 & 0.0332 \\
Octree-GS \cite{ren2024octreegs} & 0.9667 & 32.952 & 0.0335 \\
\hline
\textbf{3DGabSplat (Proposed)} & \cellcolor[HTML]{FF9999}0.9704 & \cellcolor[HTML]{FF9999}33.722 & \cellcolor[HTML]{FF9999}0.0297\\
\bottomrule
\end{tabular}
\end{table}
\begin{table*}[!t]\renewcommand{\baselinestretch}{1.0}
\renewcommand{\arraystretch}{1.0}
\setlength{\abovecaptionskip}{0pt}
\centering
\caption{Ablation study on the effect of Gabor kernel numbers per primitive on training, rendering efficiency, memory overhead, and number of primitives. 3DGS serves as the baseline for comparison. The best result in each column is highlighted in bold.}
\label{tab:ablat_freq}
\begin{tabular}{@{}lc ccccccc@{}}
\toprule
\multicolumn{2}{c}{Method} & SSIM↑ & PSNR (dB)↑ & LPIPS↓ & Train (minutes)↓ & FPS↑ & Mem (MBytes)↓ & \#Primitives (Million)↓ \\ \hline
\multicolumn{2}{@{}l}{3DGS \cite{kerbl20233dgs}} & 0.8148 & 27.408 & 0.2158 & \textbf{33} & 110 & 806 & 3.41 \\
3DGabSplat & F=1 & 0.8172 & 27.739 & 0.2114 & 34 & \textbf{134} & \textbf{700} & \textbf{2.74} \\
\textbf{3DGabSplat} & \textbf{F=2} & \textbf{0.8183} & \textbf{27.845} & \textbf{0.2103} & 36 & 132 & 732 & \textbf{2.74} \\
3DGabSplat & F=3 & 0.8170 & 27.842 & 0.2113 & 44 & 128 & 825 & 2.89 \\
3DGabSplat & F=4 & 0.8167 & 27.830 & 0.2116 & 50 & 123 & 863 & 2.86 \\
3DGabSplat & F=5 & 0.8166 & 27.827 & 0.2113 & 57 & 122 & 906 & 2.86\\
\bottomrule
\end{tabular}
\end{table*}
\begin{figure}[!t]
\renewcommand{\baselinestretch}{1.0}
\setlength{\abovecaptionskip}{0pt}
\centering
\begin{subfigure}{0.32\linewidth}
\centering
\includegraphics[width=\linewidth]{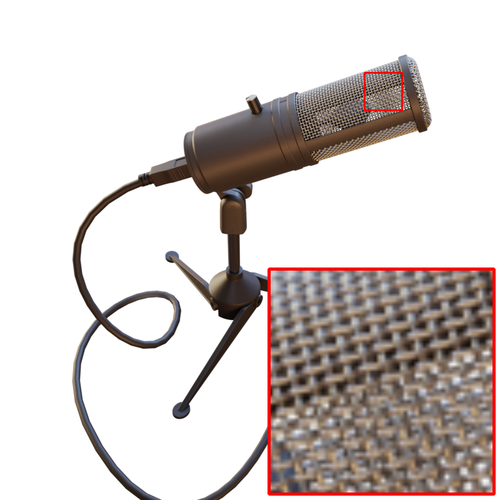}
\caption{Ground Truth}
\end{subfigure}
\begin{subfigure}{0.32\linewidth}
\centering
\includegraphics[width=\linewidth]{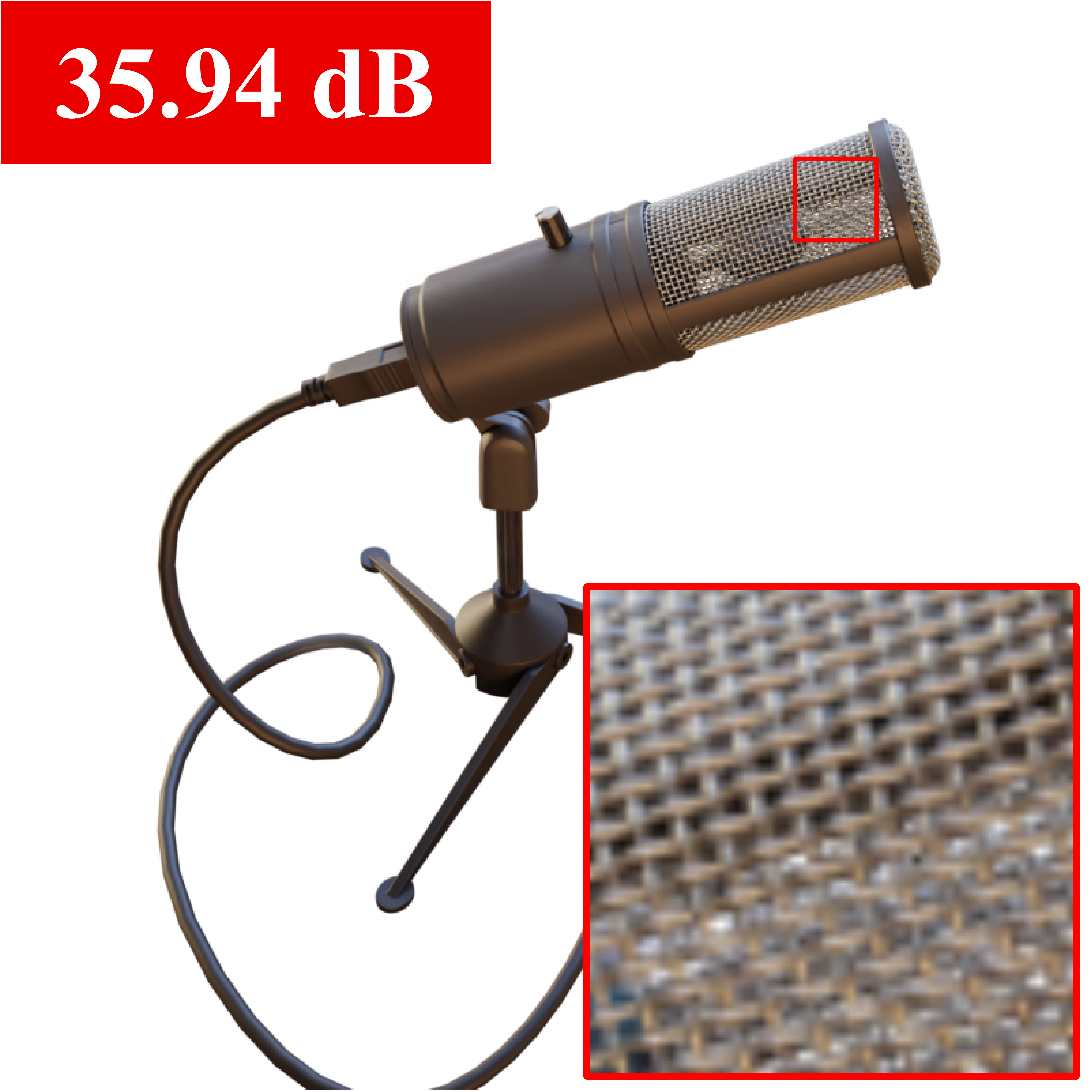}
\caption{3DGabSplat(Ours)}
\end{subfigure}
\begin{subfigure}{0.32\linewidth}
\centering
\includegraphics[width=\linewidth]{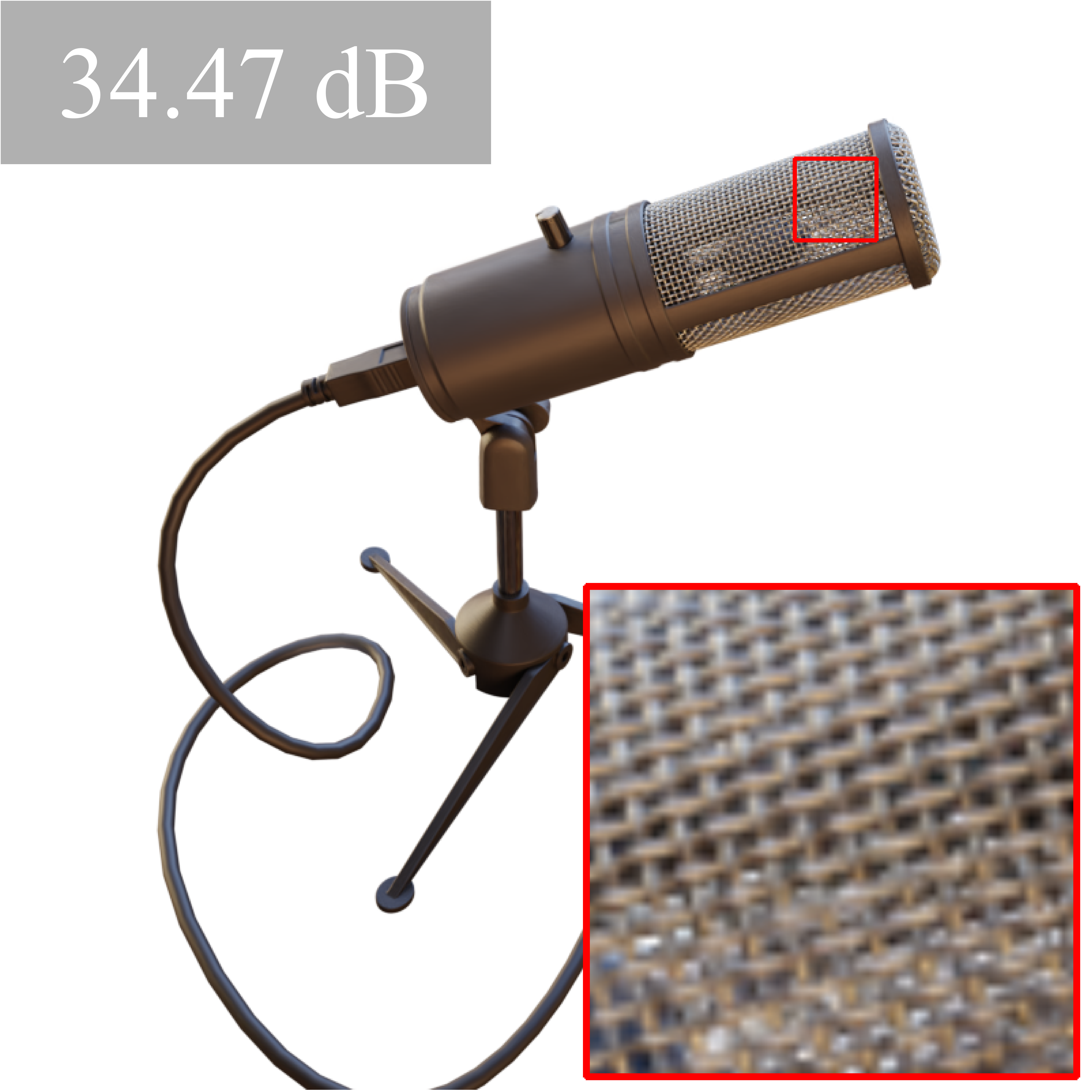}
\caption{3DGS \cite{kerbl20233dgs}}
\end{subfigure}
\caption{Qualitative comparison between our 3DGabSplat and 3DGS on the synthetic mic scene. The zoomed-in regions highlight our method’s superiority in capturing complex textures, yielding a 1.47 dB improvement in PSNR.}\label{fig:syn_comparison}
\end{figure}

Both the comparative experiments on real-world and synthetic scenes unequivocally demonstrate that our 3DGabSplat outperforms other 3DGS variants with alternative kernels, achieving state-of-the-art performance in high-fidelity novel view synthesis. 
The results further indicate the effectiveness of our 3D Gabor-based primitive as a superior alternative to the Gaussian-based primitive. By incorporating diverse frequency components for each primitive, it significantly enhances the representation of fine 3D details while enabling more efficient real-time rendering. Refer to the supplementary material for more detailed results.

\subsection{Ablation Study}
We conduct ablation studies on the key components of our method to validate their effectiveness in improving the rendering quality of 3DGabSplat for novel view synthesis. 
All ablation experiments are conducted on the Mip-NeRF 360 dataset using the same parameter configuration to ensure a fair comparison.

\noindent \textbf{Effect of the Different Frequency Numbers.}
We first analyze the impact of varying the number of Gabor kernels per primitive on rendering quality, training efficiency, inference speed, and memory consumption. The ablation results are detailed in Table \ref{tab:ablat_freq}.
Increasing the number of Gabor kernels per primitive enriches high-frequency detail representation and enhances novel view synthesis performance but comes at the cost of increased training and rendering time as well as additional memory overhead.
The optimal rendering quality is achieved at $F=2$, while further increasing the number of Gabor kernels per primitive leads to quality degradation and additional computational and storage costs. 
Therefore, to trade off rendering quality against training efficiency and storage overhead, we opt to use $F=2$ Gabor kernels per primitive in the implementation of 3DGabSplat for novel view synthesis.
Compared to 3DGS, our method achieves superior rendering performance while reducing the number of primitives for 3D representation by 20\% and increasing rendering speed by 20\% in FPS, all without significantly increasing training time.

\begin{table}[!t]
\renewcommand{\baselinestretch}{1.0}
\renewcommand{\arraystretch}{1.0}
\setlength{\abovecaptionskip}{0pt}
\centering
\caption{Ablation of key components in 3DGabSplat.}\label{tab:ablat_densify}
\resizebox{\columnwidth}{!}{%
\begin{tabular}{@{}l|ccc@{}}
\toprule
\multicolumn{1}{c|}{Method} & SSIM↑ & PSNR↑ & LPIPS↓ \\ \hline
2DGabSplat (fixed and unidirectional) & 0.8157 & 27.499 & 0.2127 \\
Proposed w/o periodic freq reset & 0.8161 & 27.793 & 0.2121 \\
Proposed w/o densification freq reset & 0.8168 & 27.819 & 0.2112 \\
Proposed w/o Gabor (degrade to Gaussian) & 0.7716 & 24.856 & 0.2573 \\
Proposed full & \textbf{0.8183} & \textbf{27.845} & \textbf{0.2103}\\
\bottomrule
\end{tabular}
}
\end{table}

\noindent \textbf{Comparison with Fixed and Undirectional Gabor Kernel.}
Additionally, prior work \cite{wurster20242dgabor} employed 2DGabSplat for image representation, utilizing fixed and unidirectional frequency components to construct Gabor kernels. 
Since this method is a special case of ours but is limited to 2D image representation, we integrate it for a fair comparison.
For implementation, we fix the frequency direction along the $x$-axis, set the maximum frequency amplitude to 12 Hz, and linearly interpolate six frequencies between zero and the maximum value. The two highest-weighted frequencies are then selected to construct the primitive.
We compare it with our 3DGabSplat using $F=2$ Gabor kernels, as shown in Table \ref{tab:ablat_freq}.
The results reveal that primitives with fixed, unidirectional frequencies are inherently limited in representing complex 3D scenes. In contrast, our method leverages 3D Gabor-based primitives with optimizable frequencies and weighting coefficients, enabling a more effective and frequency-adaptive representation of high-frequency details.

\noindent \textbf{Effect of Frequency-adaptive Optimization.}
To assess the effectiveness of our frequency-adaptive optimization method, we perform ablations on the full model by separately removing periodic frequency reset and densification frequency reset strategies. As shown in the comparison between the corresponding rows in Table \ref{tab:ablat_densify}, the frequency-adaptive optimization significantly enhances 3D rendering quality.

\noindent \textbf{Effect of Degradation to Gaussian.}
We set all the frequency components and their corresponding weighting coefficients in the trained model to zero to verify the effect of direct degradation to Gaussian-based primitives.
As shown in the fourth row of Table \ref{tab:ablat_densify}, the degradation notably impacts rendering quality, leading to a 10\% decrease in PSNR.
This further substantiates the effectiveness of our proposed Gabor-based primitives in representing high-frequency details in 3D scenes.

\section{Conclusion}
We introduce 3D Gabor Splatting (3DGabSplat), a novel approach for radiance field rendering that employs 3D Gabor-based primitives as a superior alternative to the Gaussian kernel, enabling efficient and high-fidelity novel view synthesis.
By incorporating Gabor filters with varying frequencies for each primitive and employing CUDA-based rasterizer and frequency-adaptive optimization, our method effectively captures high-frequency details in 3D scenes.
Extensive experiments demonstrate that 3DGabSplat consistently outperforms existing 3DGS variants with different kernels across both real-world and synthetic datasets.
By seamlessly integrating 3D Gabor-based primitives into 3DGS extension frameworks as a plug-and-play kernel, 3DGabSplat achieves state-of-the-art rendering performance, demonstrating its scalability and potential as a superior alternative to the Gaussian kernel for advancing 3D reconstruction research.

\begin{acks}
This work was supported in part by the National Natural Science Foundation of China under Grant 62431017, Grant 62320106003, Grant U24A20251, Grant 62125109, Grant 62371288, Grant 62301299, Grant 62401357, Grant 62401366,  Grant 62120106007, and in part by the Program of Shanghai Science and Technology Innovation Project under Grant 24BC3200800. 
\end{acks}

\bibliographystyle{ACM-Reference-Format}
\bibliography{literatures.bib}

\end{document}


\title{Supplementary Materials for ``3DGabSplat: 3D Gabor Splatting for Frequency-adaptive Radiance Field Rendering''}

\author{Junyu Zhou$^\ast$, Yuyang Huang$^\ast$, Wenrui Dai$^\dagger$, Junni Zou$^\dagger$, Ziyang Zheng, Nuowen Kan, Chenglin Li, and~Hongkai~Xiong\\
Shanghai Jiao Tong University, Shanghai, China}
\thanks{$^\ast$Both authors contributed equally to this research. $^\dagger$Corresponding authors.}

\renewcommand{\shortauthors}{Junyu Zhou et al.}

\maketitle

\appendix
\section{Detailed Derivations of 3DGabSplat}
\subsection{Projection of 3D Primitives}
In this section, we first derive the projection of our 3D Gabor-based primitives from the 3D world coordinate space to ray space.
The 3D Gabor-based primitives are composed of Gabor kernels with different frequencies, where each Gabor kernel is defined as:
\begin{equation}
g(\mathbf{x})=\exp{\left(-\frac{1}{2}(\mathbf{x}-\boldsymbol{\mu})^T\mathbf{\Sigma}^{-1}(\mathbf{x}-\boldsymbol{\mu})\right)}\cdot \cos(2\pi\boldsymbol{f}^\top (\mathbf{x}-\boldsymbol{\mu})).
\label{equ:gabor_1}
\end{equation}

Analogous to 3DGS, the projection of the Gabor kernel consists of two stages. In the first stage, coordinates of the reconstruction kernel undergo a viewing transformation from world space to camera space. In the second stage, a projective transformation transforms the coordinates from camera space to ray space. The resulting expression for the point coordinates is given by:
\begin{equation}
    \mathbf{x}''-\boldsymbol{\mu}'' \approx \mathbf{J}(\mathbf{x}'-\boldsymbol{\mu}')=\mathbf{J}\mathbf{W}(\mathbf{x}-\boldsymbol{\mu}).
\end{equation}
Here, $\mathbf{W}$ and $\mathbf{J}$ represent the world-to-camera transformation and the affine approximation Jacobian transformation matrices, respectively. 
Subsequently, we perform the inverse representation of the coordinates and substitute expression $\mathbf{x}-\boldsymbol{\mu} = (\mathbf{J}\mathbf{W})^{-1}\cdot(\mathbf{x}''-\boldsymbol{\mu}_k'')$ in \eqref{equ:gabor_1} to obtain that
\begin{equation}
g(\mathbf{x}'')=\mathcal{G}_{\boldsymbol{\mu}'', \boldsymbol{\Sigma}''}(\mathbf{x}'') \cdot\cos(2\pi{\boldsymbol{f}^{proj}}^\top (\mathbf{x}''-\boldsymbol{\mu}'')),
\end{equation}
where $\mathcal{G}_{\boldsymbol{\mu}'', \boldsymbol{\Sigma}''}(\mathbf{x}'')$ denotes the projected Gaussian kernel, and ${\boldsymbol{f}^{proj}}^\top = \boldsymbol{f}^\top(\mathbf{J}\mathbf{W})^{-1}$ represents the 3D frequency obtained after projection.

\subsection{3D-to-2D Integration}
Next, we integrate the projected 3D Gabor-based primitive along the z-axis to obtain 2D splats. For computational simplicity, we assume that the center of the Gabor kernel is located at the origin throughout the derivation, leading to the following expression:
\begin{equation}
g(\mathbf{x})=\exp{\left(-\frac{1}{2}\mathbf{x}^T\mathbf{\Sigma}^{-1}\mathbf{x}\right)}\cdot \cos(2\pi\boldsymbol{f}^\top \mathbf{x}).
\label{equ:gabor_2}
\end{equation}

Given that the inverse of the covariance matrix is represented as
\begin{equation}
\boldsymbol{\Sigma}^{-1}=\begin{pmatrix}S_{00}&S_{01}&S_{02}\\S_{10}&S_{11}&S_{12}\\S_{20}&S_{21}&S_{22}\end{pmatrix}.
\end{equation}
We can further simplify \eqref{equ:gabor_2} as follows:
\begin{align}
g(x,y,z)\!=& \exp{\!\left(-\frac{1}{2}(S_{00}x^2\!+\!S_{11}y^2\!+\!S_{22}z^2\!+\!2S_{01}xy\!+\!2S_{02}xz\!+\!2S_{12}yz)\right)}\cdot \nonumber\\
&\cos(2\pi(f_x x + f_y y + f_z z)) \nonumber\\
=&\exp{\!\left(-\frac{1}{2}(S_{00}x^2\!+\!S_{11}y^2\!+\!S_{22}z^2 \!+\!2S_{01}xy \!+\!2S_{02}xz \!+\! 2S_{12}yz)\right) }\cdot \nonumber\\
&\frac{1}{2}(\exp{\!(j2\pi(f_x x\!+\!f_y y\!+\!f_z z))} \!+\! \exp{\!(-j2\pi(f_x x\!+\!f_y y \!+\! f_z z))}).
\end{align}

In the subsequent integration process, we decompose $g(x,y,z)$ into two parts, \emph{i.e.}, $g(x,y,z) = \frac{1}{2}[g_1(x,y,z) + g_2(x,y,z)]$, which can be expressed as:
\begin{align}
g_1(x,y,z)\!=\! \exp\bigg(&\!-\!\frac{1}{2}(S_{00}x^2\!+\!S_{11}y^2\!+\!S_{22}z^2\!+\!2S_{01}xy\!+\!2S_{02}xz\!+\!2S_{12}yz)\nonumber\\
 &+ j2\pi(f_x x + f_y y + f_z z)\bigg),
\end{align}
and
\begin{align}
g_2(x,y,z)\!=\!\exp\bigg(&\!-\frac{1}{2}(S_{00}x^2\!+\!S_{11}y^2\!+\!S_{22}z^2\!+\!2S_{01}xy\!+\!2S_{02}xz\!+\!2S_{12}yz)\nonumber\\
&- j2\pi(f_x x + f_y y + f_z z)\bigg).
\end{align}

Integrating the first term $g_1(x,y,z)$ along the z-axis, we obtain:
\begin{align}\label{equ:g_1}
&\int g_1(x,y,z) \mathrm{d}z \nonumber\\
&\quad =\int \exp\bigg(-\frac{1}{2}(S_{00}x^2\!+\!S_{11}y^2\!+\!S_{22}z^2\!+\!2S_{01}xy\!+\!2S_{02}xz\!+\!2S_{12}yz)\nonumber\\
&\qquad\qquad\qquad + j2\pi(f_x x + f_y y + f_z z)\bigg) \mathrm{d}z\nonumber\\
&\quad= \exp{\left(-\frac{1}{2}S_{00}x^2-\frac{1}{2}S_{11}y^2-S_{01}xy+j2\pi(f_x x + f_y y)\right)}\cdot \nonumber\\
&\qquad\int\exp{\left(-\frac{1}{2}S_{22}z^2-(S_{02}x+S_{12}y-j2\pi f_z)z\right)}\mathrm{d}z.
\end{align}
In \eqref{equ:g_1}, we consider the term $\int\!\exp(-\frac{1}{2}S_{22}z^2-(S_{02}x+S_{12}y-j2\pi f_z)z)\mathrm{d}z$.
\begin{align}\label{equ:g_1-1}
& \int\exp{\left(-\frac{1}{2}S_{22}z^2-(S_{02}x+S_{12}y-j2\pi f_z)z\right)}\mathrm{d}z\nonumber\\
&\quad = \int\exp\bigg(-\frac{1}{2}S_{22}\left(z+\frac{S_{02}x+S_{12}y-j2\pi f_z}{S_{22}}\right)^2\nonumber\\ &\qquad\qquad+\frac{(S_{02}x+S_{12}y-j2\pi f_z)^2}{2S_{22}}\bigg)\mathrm{d}z\nonumber\\
&\quad =\exp{\left(\frac{(S_{02}x+S_{12}y-j2\pi f_z)^2}{2S_{22}}\right)}\cdot \nonumber\\
&\qquad\qquad \int\exp{\left(-\frac{1}{2}S_{22}(z+\frac{S_{02}x+S_{12}y-j2\pi f_z}{S_{22}})^2\right)}\mathrm{d}z.
\end{align}
From \eqref{equ:g_1} and \eqref{equ:g_1-1}, we find that 
\begin{align}\label{equ:g_1-2}
&\int\!g_1(x,y,z)\mathrm{d}z\nonumber\\
&\quad=C_1\exp\!\bigg(-\frac{1}{2}S_{00}x^2\!-\!\frac{1}{2}S_{11}y^2\!-\!S_{01}xy\!+\!j2\pi(f_x x\!+\!f_y y)
\nonumber\\
&\quad\qquad\qquad+\frac{(S_{02}x\!+\!S_{12}y\!-\!j2\pi f_z)^2}{2S_{22}}\bigg)\nonumber\\
&\quad= C_1\exp\!\bigg(-\!\frac{1}{2}S_{00}x^2\!-\!\frac{1}{2}S_{11}y^2\!-\!S_{01}xy\!+\!\frac{(S_{02}x\!+\!S_{12}y)^2}{2S_{22}} \nonumber\\
&\quad\qquad\qquad+\!j2\pi\bigg(f_x x\!+\!f_y y\!-\!\frac{(S_{02}x\!+\!S_{12}y)f_z}{S_{22}}\bigg)\!-\!\frac{(j2\pi f_z)^2}{2S_{22}}\bigg).
\end{align}
Therefore, we obtain that 
\begin{align}
\!&\int g_1(x,y,z)\mathrm{d}z\nonumber\\
&\quad= C_1C_2 \cdot \mathcal{G}^{2D}(x,y)\exp\!\left(j2\pi\left(\left(f_x\!-\!\frac{S_{02}}{S_{22}}f_z\right)x\!+\!\left(f_y\!-\!\frac{S_{12}}{S_{22}}f_z\right)y\right)\right),
\end{align}
where $C_1 = (2S_{22}\pi)^{-1/2}$ and $C_2 = \exp{((2\pi f_z)^2/(2S_{22}))}$ represent the coefficients of the exponential term.

Similarly, we apply the same integration procedure to $g_2(x,y,z)$, and obtain that 
\begin{align}\label{equ:g_2}
&\int g_2(x,y,z)\mathrm{d}z \nonumber\\ 
&\quad= C_1C_2\cdot\mathcal{G}^{2D}(x,y)\exp\!\left(-j2\pi\left(\left(f_x\!-\!\frac{S_{02}}{S_{22}}f_z\right)x\!+\!\left(f_y\!-\!\frac{S_{12}}{S_{22}}f_z\right)y\right)\right).
\end{align}
Combining \eqref{equ:g_1-2} and \eqref{equ:g_2}, we obtian the total integration result that
\begin{align}
\hat{g}(x,y)&=\int g(x,y,z)dz=\frac{1}{2}\left(\int g_1(x,y,z)\mathrm{d}z+\int g_2(x,y,z)\mathrm{d}z\right)\nonumber\\
&=\mathcal{G}^{2D}(x,y)\cdot\cos(2\pi(\hat{f}_xx+\hat{f}_yy)),
\end{align}
where the 2D frequency obtained by the integration is 
\begin{equation}\left\{\begin{aligned}
{\hat{f}_{x}} &= {f}_x^{proj} - \frac{S_{02}}{S_{22}}\cdot{f}_{z}^{proj}\\
{\hat{f}_{y}} &= {f}_{y}^{proj} - \frac{S_{12}}{S_{22}}\cdot{f}_{z}^{proj}
\end{aligned}\right..
\end{equation}

\begin{table*}[!p]
\setlength{\tabcolsep}{1pt}
\centering
\caption{Efficiency evaluation of 3DGabSplat and baselines on MipNeRF360, TNT, and DB datasets.}
\label{tab:efficiency1}
\begin{tabular}{@{}l ccccc ccccc ccccc@{}}
\toprule
Dataset & \multicolumn{5}{c}{Mip-NeRF360} & \multicolumn{5}{c}{Tanks\&Temples} & \multicolumn{5}{c}{DeepBlending} \\
\cmidrule{1-1}\cmidrule(lr){2-6}\cmidrule(lr){7-11}\cmidrule(lr){12-16}
\multirow{2}*{Method | Metric} & PSNR↑ & Train↓ & FPS↑ & Mem↓ & \# Prim↓ & PSNR↑ & Train↓ & FPS↑ & Mem↓ & \# Prim↓ & PSNR↑ & Train↓ & FPS↑ & Mem↓ & \# Prim↓ \\ 
& (dB) & (min) & & (MB) & (Million) & (dB) & (min) & & (MB) & (Million) & (dB) & (min) & & (MB) & (Million)\\
\midrule
3DGS & 27.41 & 33 & 110 & 806 & 3.41 & 23.69 & 20 & 176 & 442 & 1.87 & 29.55 & 29 & 122 & 666 & 2.82 \\
\textbf{Proposed} & 27.85 & 36 & 132 & 732 & 2.74 & 24.49 & 22 & 194 & 371 & 1.37 & 30.09 & 33 & 167 & 538 & 1.98 \\ \midrule
Scaffold-GS & 27.72 & 26 & 191 & 180 & 0.55 & 24.00 & 17 & 213 & 77 & 0.46 & 30.21 & 20 & 357 & 54 & 0.09 \\
\makecell[l]{\textbf{Proposed}\\ \textbf{ +Scaffold-GS}} & 27.84 & 29 & 214 & 192 & 0.56 & 24.52 & 20 & 239 & 84 & 0.46 & 30.52 & 23 & 391 & 43 & 0.06 \\ \midrule
Mip-Splat+AbsGS & 27.66 & 43 & 88 & 1029 & 4.28 & 23.71 & 25 & 135 & 572 & 2.38 & 29.31 & 38 & 97 & 822 & 3.42 \\
\makecell[l]{\textbf{Proposed}\\ \textbf{ +Mip-Splat+AbsGS}} & 27.95 & 48 & 107 & 966 & 3.68 & 24.37 & 31 & 147 & 523 & 1.93 & 29.79 & 43 & 113 & 724 & 2.67 \\ \bottomrule
\end{tabular}
\vspace{6pt}
\centering
\caption{Efficiency evaluation of 3DGabSplat and baselines on BungeeNeRF, and OMMO datasets.}\label{tab:efficiency2}
\begin{tabular}{@{}l ccccccc@{}}
\toprule
Dataset & \multicolumn{7}{c}{BungeeNeRF}\\
\midrule
Method | Metric & SSIM↑ & PSNR (dB)↑ & LPIPS↓ & Train (minutes)↓ & FPS↑ & Mem (MBytes)↓ & \# Prim (Million)↓ \\ 
\midrule
3DGS & 0.893 & 26.60 & 0.130 & 70 & 45 & 1832 & 7.75 \\
\textbf{Proposed} & 0.901 & 26.94 & 0.124 & 78 & 55 & 1845 & 6.84 \\ 
\midrule
Scaffold-GS & 0.877 & 26.29 & 0.158 & 33 & 105 & 126 & 1.91 \\
\makecell[l]{\textbf{Proposed}\\ \textbf{ +Scaffold-GS}} & 0.882 & 26.52 & 0.155 & 36 & 118 & 138 & 1.88 \\ 
\midrule
Mip-Splat+AbsGS & 0.909 & 27.40 & 0.113 & 90 & 20 & 2140 & 8.90 \\
\makecell[l]{\textbf{Proposed}\\ \textbf{ +Mip-Splat+AbsGS}} & 0.913 & 27.65 & 0.108 & 99 & 27 & 2178 & 8.07 \\ 
\bottomrule
Dataset & \multicolumn{7}{c}{OMMO} \\
\midrule
Method | Metric & SSIM↑ & PSNR (dB)↑ & LPIPS↓ & Train (minutes)↓ & FPS↑ & Mem (MBytes)↓ & \# Prim (Million)↓ \\ 
\midrule
3DGS & 0.831 & 25.87 & 0.215 & 29 & 130 & 628 & 2.65 \\
\textbf{Proposed} & 0.856 & 27.14 & 0.187 & 32 & 149 & 593 & 2.19 \\ \midrule
Scaffold-GS & 0.839 & 26.55 & 0.201 & 25 & 118 & 198 & 0.67 \\
\makecell[l]{\textbf{Proposed}\\ \textbf{ +Scaffold-GS}} & 0.849 & 27.06 & 0.189 & 28 & 111 & 214 & 0.67 \\ \midrule
Mip-Splat+AbsGS & 0.841 & 26.26 & 0.200 & 37 & 104 & 806 & 3.36 \\
\makecell[l]{\textbf{Proposed}\\ \textbf{ +Mip-Splat+AbsGS}} & 0.863 & 27.29 & 0.176 & 43 & 118 & 696 & 2.75 \\ 
\bottomrule
\end{tabular}
\vspace{6pt}
\centering
\caption{Comparison with recent state-of-the-art methods on the MipNeRF360 dataset.}
\label{tab:new_sota}
\begin{tabular}{@{}l ccccccc@{}}
\toprule
Method & SSIM↑ & PSNR (dB)↑ & LPIPS↓ & Train (minutes)↓ & FPS↑ & Mem (MBytes)↓ & \# Prim (Million)↓ \\ \midrule
3DGS & 0.814 & 27.41 & 0.216 & \textbf{33} & 110 & 806 & 3.41 \\
3DHGS & 0.813 & 27.71 & 0.215 & 40 & 129 & 773 & 3.22 \\
\textbf{Proposed} & \textbf{0.818} & \textbf{27.85} & \textbf{0.210} & 36 & \textbf{132} & \textbf{732} & \textbf{2.74} \\ \midrule
3DGS-MCMC & 0.834 & 28.01 & 0.194 & \textbf{50} & 133 & 732 & 3.10 \\
DBS & 0.833 & 28.36 & 0.196 & 57 & \textbf{62} & \textbf{355} & 3.10 \\
\textbf{Proposed+MCMC} & \textbf{0.839} & \textbf{28.46} & \textbf{0.181} & 59 & 124 & 826 & 3.10 \\ \bottomrule
\end{tabular}
\end{table*}
\begin{figure*}[!p]
    \centering
    \begin{subfigure}[b]{0.19\textwidth}
        \captionsetup{labelformat=empty}
        \caption{Ground Truth}
        \centering
        \includegraphics[width=\textwidth]{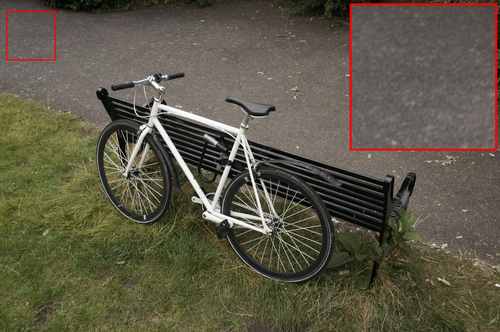}
    \end{subfigure}
    \begin{subfigure}[b]{0.19\textwidth}
        \captionsetup{labelformat=empty}
        \caption{3DGabSplat (Ours)}
        \centering
        \includegraphics[width=\textwidth]{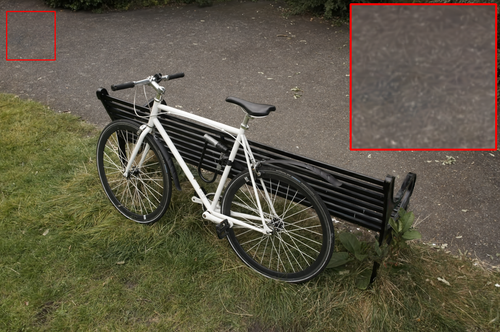}
    \end{subfigure}
    \begin{subfigure}[b]{0.19\textwidth}
        \captionsetup{labelformat=empty}
        \caption{3DGS}
        \centering
        \includegraphics[width=\textwidth]{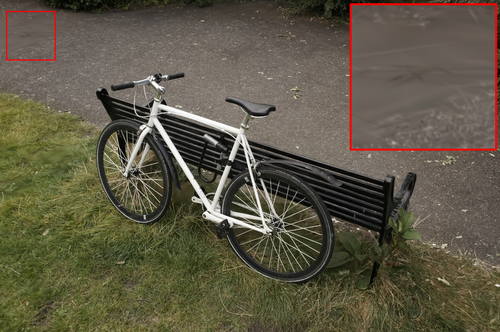}
    \end{subfigure}
    \begin{subfigure}[b]{0.19\textwidth}
        \captionsetup{labelformat=empty}
        \caption{GES}
        \centering
        \includegraphics[width=\textwidth]{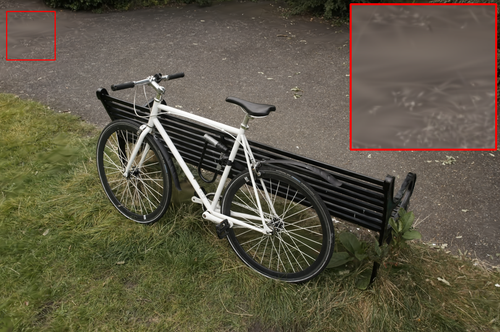}
    \end{subfigure}
    \begin{subfigure}[b]{0.19\textwidth}
        \captionsetup{labelformat=empty}
        \caption{3DCS}
        \centering
        \includegraphics[width=\textwidth]{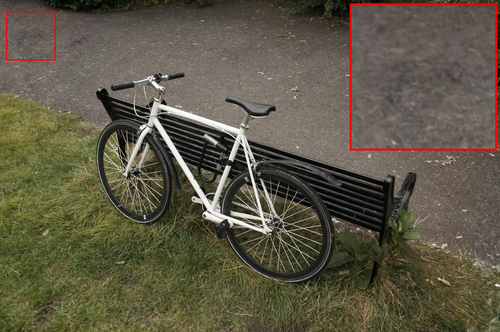}
    \end{subfigure}

\vspace{1mm} %

\begin{subfigure}[b]{0.19\textwidth}
        \captionsetup{labelformat=empty}
        \centering
        \includegraphics[width=\textwidth]{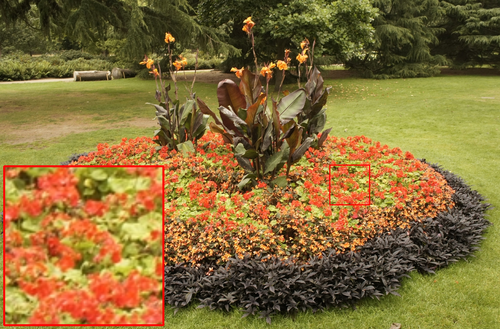}
    \end{subfigure}
    \begin{subfigure}[b]{0.19\textwidth}
        \captionsetup{labelformat=empty}
        \centering
        \includegraphics[width=\textwidth]{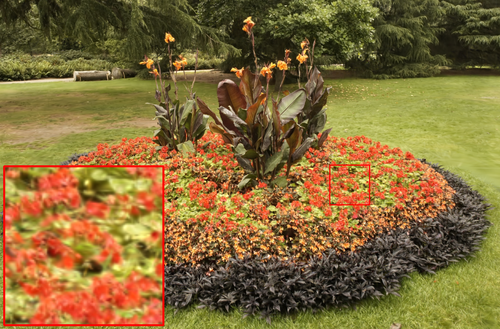}
    \end{subfigure}
    \begin{subfigure}[b]{0.19\textwidth}
        \captionsetup{labelformat=empty}
        \centering
        \includegraphics[width=\textwidth]{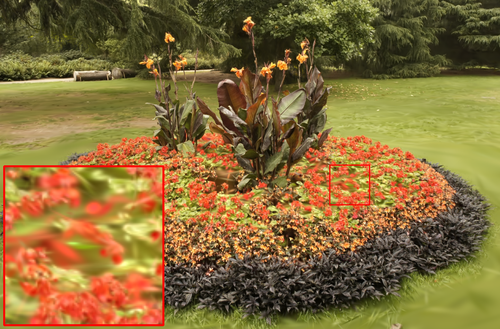}
    \end{subfigure}
    \begin{subfigure}[b]{0.19\textwidth}
        \captionsetup{labelformat=empty}
        \centering
        \includegraphics[width=\textwidth]{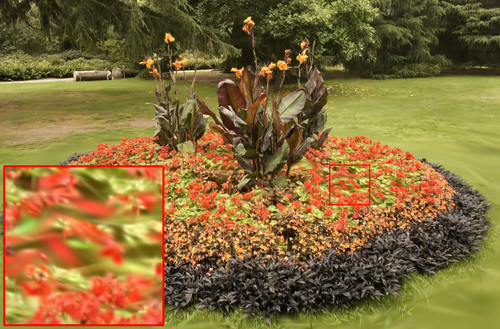}
    \end{subfigure}
    \begin{subfigure}[b]{0.19\textwidth}
        \captionsetup{labelformat=empty}
        \centering
        \includegraphics[width=\textwidth]{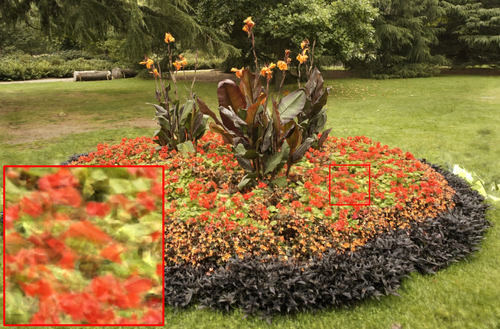}
    \end{subfigure}

\vspace{1mm} %

\begin{subfigure}[b]{0.19\textwidth}
        \captionsetup{labelformat=empty}
        \centering
        \includegraphics[width=\textwidth]{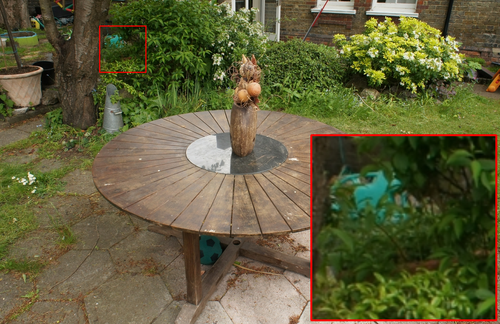}
    \end{subfigure}
    \begin{subfigure}[b]{0.19\textwidth}
        \captionsetup{labelformat=empty}
        \centering
        \includegraphics[width=\textwidth]{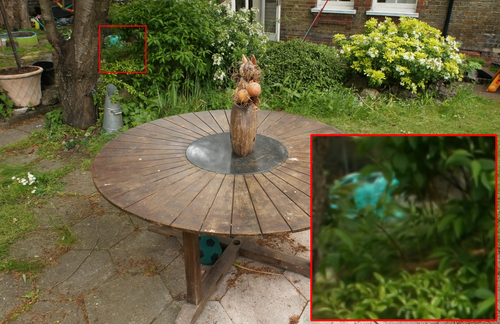}
    \end{subfigure}
    \begin{subfigure}[b]{0.19\textwidth}
        \captionsetup{labelformat=empty}
        \centering
        \includegraphics[width=\textwidth]{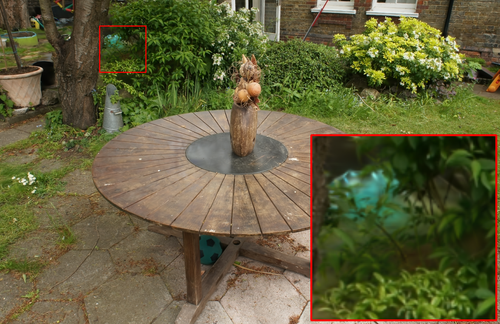}
    \end{subfigure}
    \begin{subfigure}[b]{0.19\textwidth}
        \captionsetup{labelformat=empty}
        \centering
        \includegraphics[width=\textwidth]{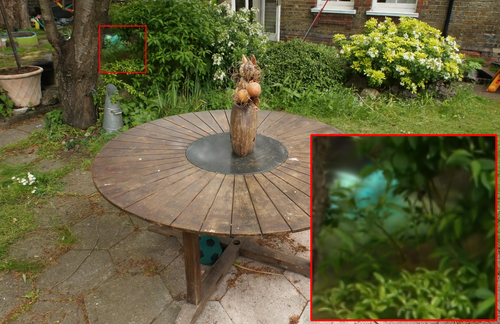}
    \end{subfigure}
    \begin{subfigure}[b]{0.19\textwidth}
        \captionsetup{labelformat=empty}
        \centering
        \includegraphics[width=\textwidth]{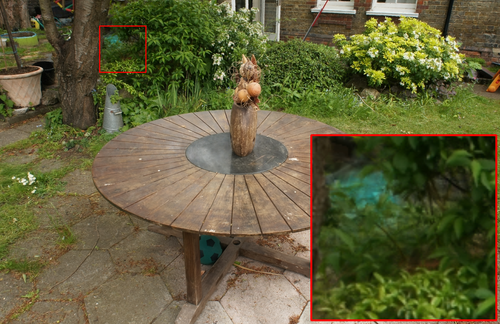}
    \end{subfigure}

\vspace{1mm} %

\begin{subfigure}[b]{0.19\textwidth}
        \captionsetup{labelformat=empty}
        \centering
        \includegraphics[width=\textwidth]{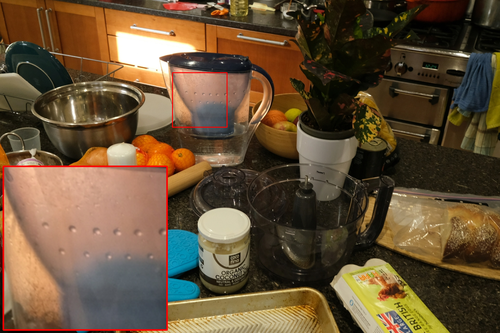}
    \end{subfigure}
    \begin{subfigure}[b]{0.19\textwidth}
        \captionsetup{labelformat=empty}
        \centering
        \includegraphics[width=\textwidth]{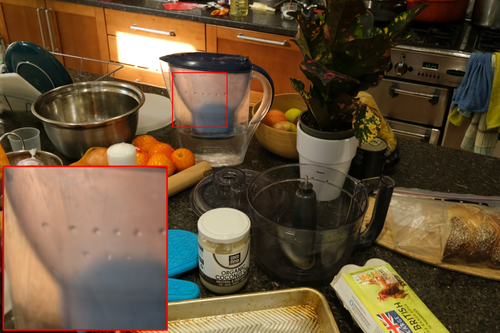}
    \end{subfigure}
    \begin{subfigure}[b]{0.19\textwidth}
        \captionsetup{labelformat=empty}
        \centering
        \includegraphics[width=\textwidth]{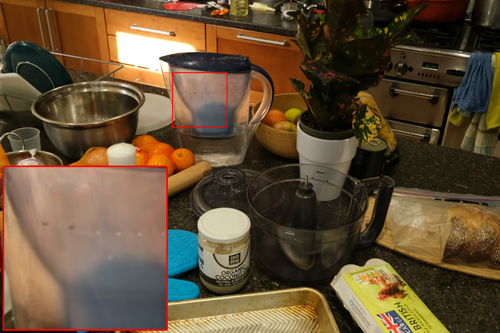}
    \end{subfigure}
    \begin{subfigure}[b]{0.19\textwidth}
        \captionsetup{labelformat=empty}
        \centering
        \includegraphics[width=\textwidth]{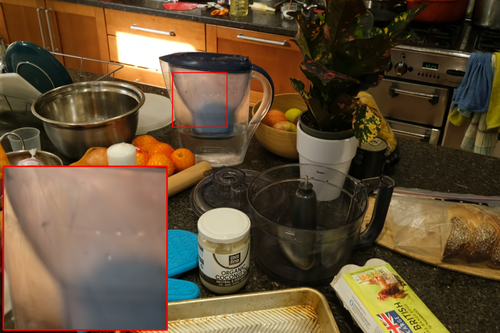}
    \end{subfigure}
    \begin{subfigure}[b]{0.19\textwidth}
        \captionsetup{labelformat=empty}
        \centering
        \includegraphics[width=\textwidth]{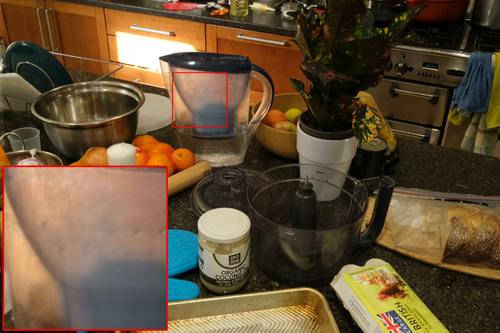}
    \end{subfigure}
\vspace{1mm} %
\begin{subfigure}[b]{0.19\textwidth}
        \captionsetup{labelformat=empty}
        \centering
        \includegraphics[width=\textwidth]{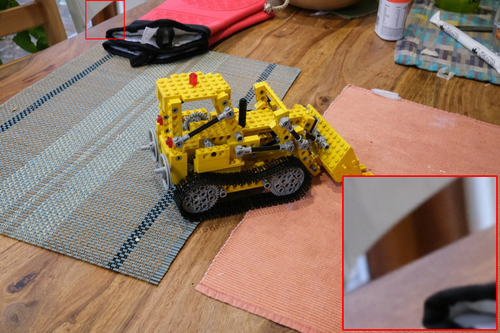}
    \end{subfigure}
    \begin{subfigure}[b]{0.19\textwidth}
        \captionsetup{labelformat=empty}
        \centering
        \includegraphics[width=\textwidth]{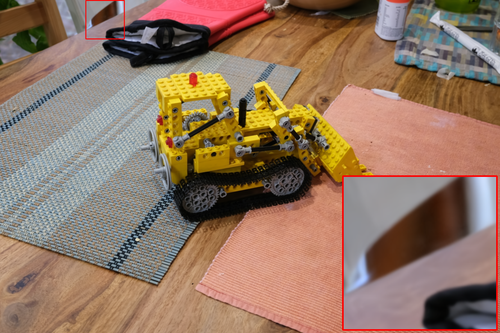}
    \end{subfigure}
    \begin{subfigure}[b]{0.19\textwidth}
        \captionsetup{labelformat=empty}
        \centering
        \includegraphics[width=\textwidth]{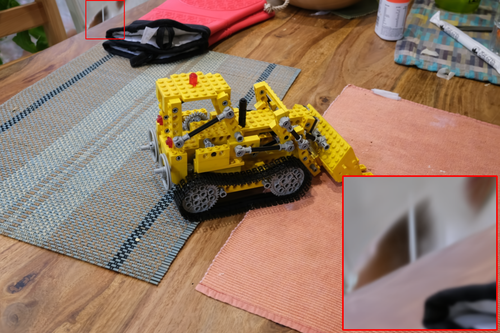}
    \end{subfigure}
    \begin{subfigure}[b]{0.19\textwidth}
        \captionsetup{labelformat=empty}
        \centering
        \includegraphics[width=\textwidth]{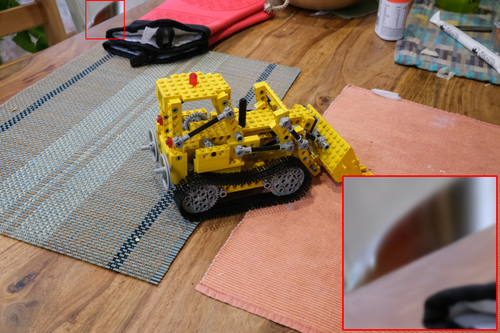}
    \end{subfigure}
    \begin{subfigure}[b]{0.19\textwidth}
        \captionsetup{labelformat=empty}
        \centering
        \includegraphics[width=\textwidth]{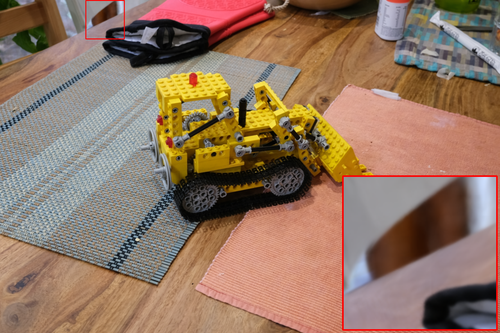}
    \end{subfigure}
\vspace{1mm} %
\begin{subfigure}[b]{0.19\textwidth}
        \captionsetup{labelformat=empty}
        \centering
        \includegraphics[width=\textwidth]{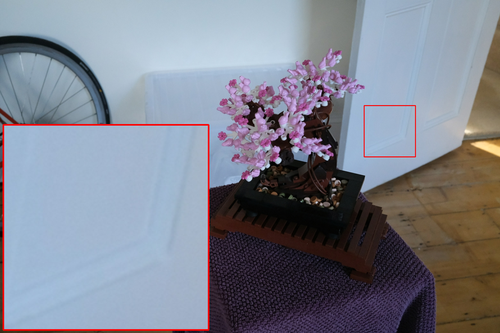}
    \end{subfigure}
    \begin{subfigure}[b]{0.19\textwidth}
        \captionsetup{labelformat=empty}
        \centering
        \includegraphics[width=\textwidth]{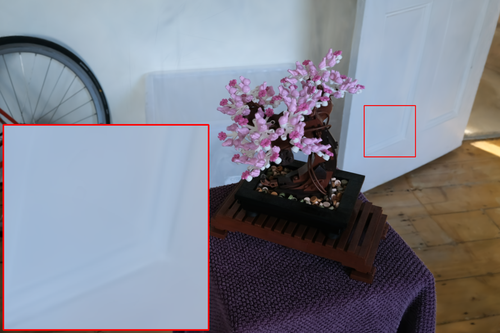}
    \end{subfigure}
    \begin{subfigure}[b]{0.19\textwidth}
        \captionsetup{labelformat=empty}
        \centering
        \includegraphics[width=\textwidth]{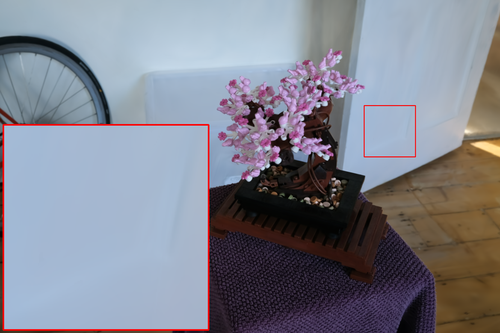}
    \end{subfigure}
    \begin{subfigure}[b]{0.19\textwidth}
        \captionsetup{labelformat=empty}
        \centering
        \includegraphics[width=\textwidth]{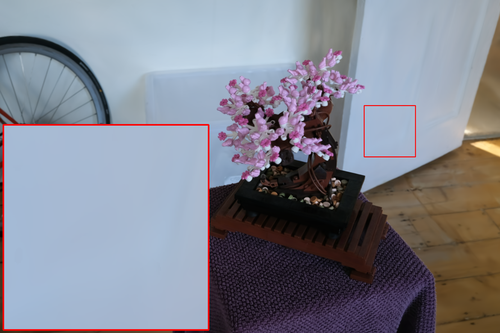}
    \end{subfigure}
    \begin{subfigure}[b]{0.19\textwidth}
        \captionsetup{labelformat=empty}
        \centering
        \includegraphics[width=\textwidth]{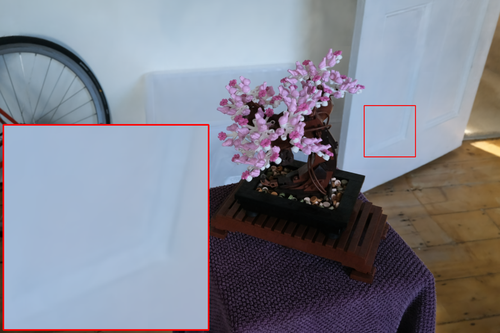}
    \end{subfigure}
\vspace{1mm} %
\begin{subfigure}[b]{0.19\textwidth}
        \captionsetup{labelformat=empty}
        \centering
        \includegraphics[width=\textwidth]{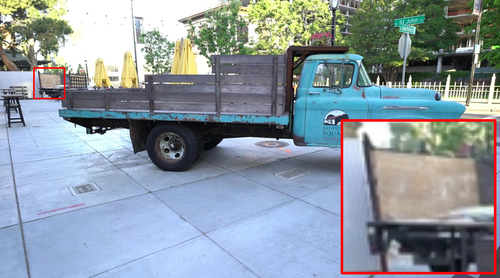}
    \end{subfigure}
    \begin{subfigure}[b]{0.19\textwidth}
        \captionsetup{labelformat=empty}
        \centering
        \includegraphics[width=\textwidth]{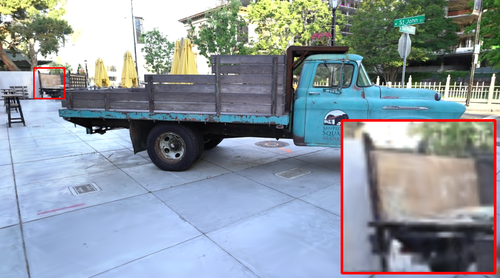}
    \end{subfigure}
    \begin{subfigure}[b]{0.19\textwidth}
        \captionsetup{labelformat=empty}
        \centering
        \includegraphics[width=\textwidth]{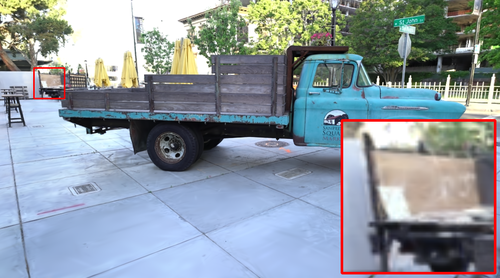}
    \end{subfigure}
    \begin{subfigure}[b]{0.19\textwidth}
        \captionsetup{labelformat=empty}
        \centering
        \includegraphics[width=\textwidth]{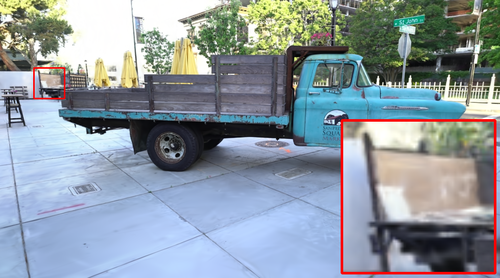}
    \end{subfigure}
    \begin{subfigure}[b]{0.19\textwidth}
        \captionsetup{labelformat=empty}
        \centering
        \includegraphics[width=\textwidth]{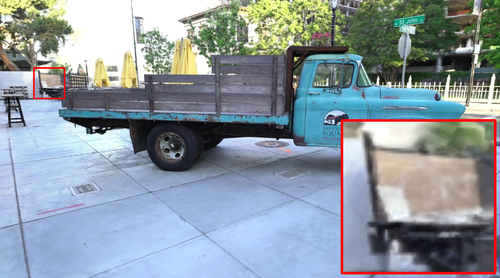}
    \end{subfigure}
\vspace{1mm} %
\begin{subfigure}[b]{0.19\textwidth}
        \captionsetup{labelformat=empty}
        \centering
        \includegraphics[width=\textwidth]{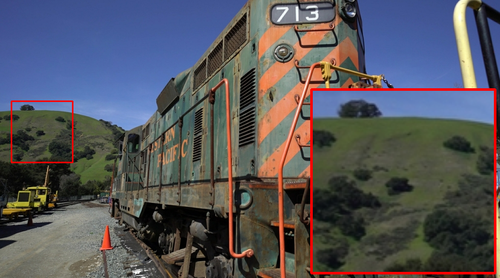}
    \end{subfigure}
    \begin{subfigure}[b]{0.19\textwidth}
        \captionsetup{labelformat=empty}
        \centering
        \includegraphics[width=\textwidth]{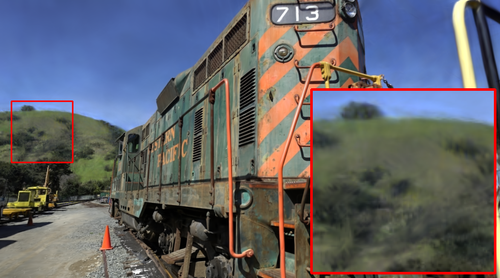}
    \end{subfigure}
    \begin{subfigure}[b]{0.19\textwidth}
        \captionsetup{labelformat=empty}
        \centering
        \includegraphics[width=\textwidth]{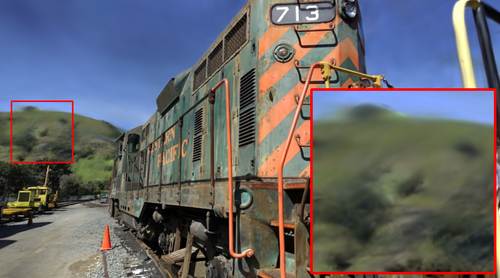}
    \end{subfigure}
    \begin{subfigure}[b]{0.19\textwidth}
        \captionsetup{labelformat=empty}
        \centering
        \includegraphics[width=\textwidth]{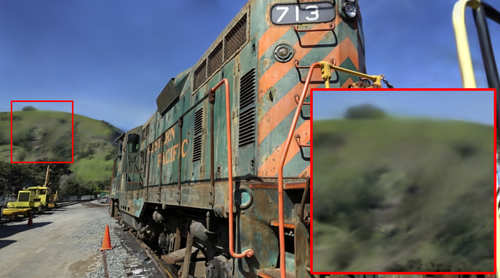}
    \end{subfigure}
    \begin{subfigure}[b]{0.19\textwidth}
        \captionsetup{labelformat=empty}
        \centering
        \includegraphics[width=\textwidth]{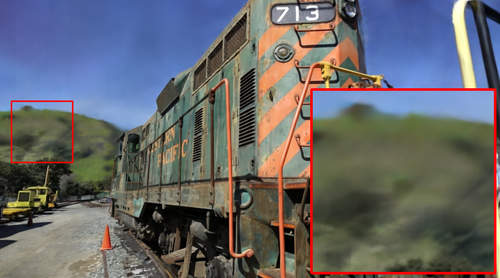}
    \end{subfigure}
\vspace{1mm} %
\begin{subfigure}[b]{0.19\textwidth}
\captionsetup{labelformat=empty}
\centering
\includegraphics[width=\textwidth]{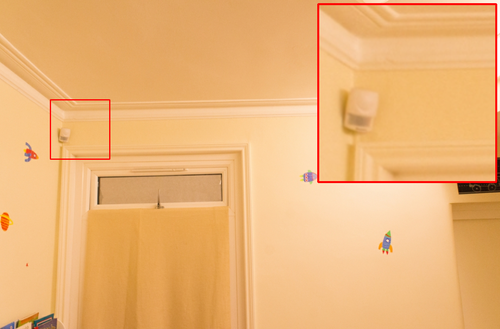}
\end{subfigure}
\begin{subfigure}[b]{0.19\textwidth}
\captionsetup{labelformat=empty}
\centering
\includegraphics[width=\textwidth]{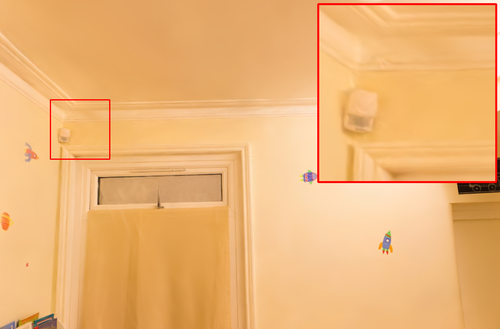}
\end{subfigure}
\begin{subfigure}[b]{0.19\textwidth}
\captionsetup{labelformat=empty}
\centering
\includegraphics[width=\textwidth]{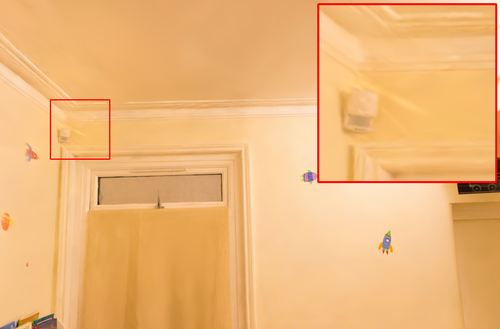}
\end{subfigure}
\begin{subfigure}[b]{0.19\textwidth}
\captionsetup{labelformat=empty}
\centering
\includegraphics[width=\textwidth]{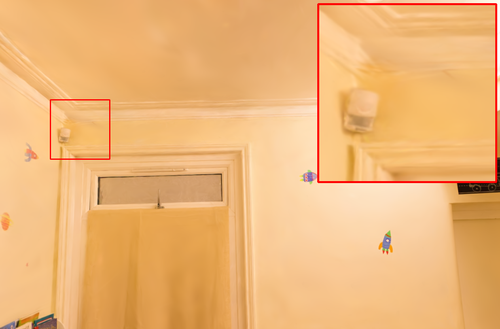}
\end{subfigure}
\begin{subfigure}[b]{0.19\textwidth}
\captionsetup{labelformat=empty}
\centering
\includegraphics[width=\textwidth]{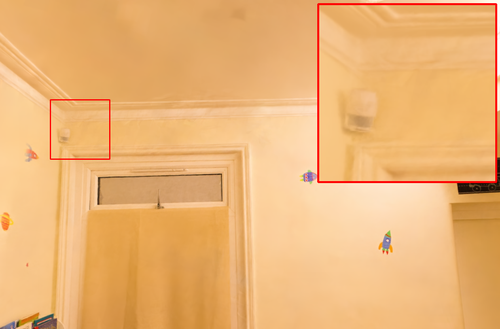}
\end{subfigure}
\caption{Additional visual comparisons of novel view synthesis results between our proposed 3DGaborSplat and existing baselines across different scenes.}\label{fig:supple_visualization}
\end{figure*}
\begin{table*}[!p]
\setlength{\tabcolsep}{4pt}
\centering
\caption{Per-scene SSIM comparison on Mip-NeRF360 Dataset.}
\label{tab:mip360-ssim}
\begin{tabular}{@{}l ccccc cccc ccc@{}}
\toprule
\multirow{2}*{Method} & \multicolumn{5}{c}{Outdoor} & \multicolumn{4}{c}{Indoor} & \multirow{2}*{Average} & \multirow{2}*{Outdoor} & \multirow{2}*{Indoor} \\
\cmidrule(lr){2-6}\cmidrule(lr){7-10}
& bicycle & flowers & garden & stump & treehill & room & counter & kitchen & bonsai & & & \\ 
\midrule
Mip-NeRF360 & 0.6850 & 0.5830 & 0.8130 & 0.7440 & 0.6320 & 0.9130 & 0.8940 & 0.9200 & 0.9410 & 0.7917 & 0.6914 & 0.9170 \\
3DGS & 0.7710 & 0.6050 & 0.8680 & 0.7750 & 0.6380 & 0.9140 & 0.9050 & 0.9220 & 0.9380 & 0.8151 & 0.7314 & 0.9198 \\
3DGS* & 0.7640 & 0.6044 & 0.8667 & 0.7715 & 0.6318 & 0.9191 & 0.9084 & 0.9267 & 0.9410 & 0.8148 & 0.7277 & 0.9238 \\
2DGS & 0.7310 & 0.5730 & 0.8450 & 0.7640 & 0.6300 & 0.9180 & 0.9080 & 0.9270 & 0.9400 & 0.8040 & 0.7086 & 0.9233 \\
GES-densify\_grad\_th\_3e-4 & 0.7305 & 0.5572 & 0.8507 & 0.7435 & 0.6171 & 0.9128 & 0.9020 & 0.9222 & 0.9368 & 0.7970 & 0.6998 & 0.9184 \\
GES-densify\_grad\_th\_2e-4 & 0.7639 & 0.6037 & 0.8655 & 0.7704 & 0.6319 & 0.9191 & 0.9083 & 0.9270 & 0.9419 & 0.8146 & 0.7271 & 0.9241 \\
3DCS & 0.7370 & 0.5750 & 0.8500 & 0.7460 & 0.5950 & 0.9250 & 0.9090 & 0.9300 & 0.9450 & 0.8013 & 0.7006 & 0.9273 \\
\makecell[l]{\textbf{3DGaborSplat}\\ \textbf{(Proposed)}} & 0.7680 & 0.6089 & 0.8703 & 0.7689 & 0.6324 & 0.9253 & 0.9136 & 0.9306 & 0.9467 & 0.8183 & 0.7297 & 0.9291 \\
\midrule
Scaffold-GS & 0.7390 & 0.5770 & 0.8510 & 0.7640 & 0.6410 & \cellcolor[HTML]{FFF6B2}0.9290 & \cellcolor[HTML]{FFCC99}0.9170 & \cellcolor[HTML]{FFCC99}0.9320 & \cellcolor[HTML]{FFF6B2}0.9490 & 0.8110 & 0.7144 & \cellcolor[HTML]{FFF6B2}0.9318 \\
Scaffold-GS* & 0.7593 & 0.5949 & 0.8628 & 0.7667 & \cellcolor[HTML]{FFF6B2}0.6436 & 0.9239 & 0.9116 & 0.9276 & 0.9438 & 0.8149 & 0.7254 & 0.9267 \\
Octree-GS & 0.7460 & 0.5860 & 0.8570 & 0.7600 & \cellcolor[HTML]{FF9999}0.6460 & \cellcolor[HTML]{FF9999}0.9340 & \cellcolor[HTML]{FF9999}0.9240 & \cellcolor[HTML]{FFCC99}0.9320 & \cellcolor[HTML]{FF9999}0.9520 & 0.8152 & 0.7190 & \cellcolor[HTML]{FF9999}0.9355 \\
Octree-GS* & 0.7621 & 0.6009 & 0.8665 & 0.7645 & \cellcolor[HTML]{FF9999}0.6460 & 0.9250 & 0.9078 & 0.9200 & 0.9292 & 0.8136 & 0.7280 & 0.9205 \\
\makecell[l]{\textbf{3DGaborSplat}\\ \textbf{ +Scaffold-GS}} & 0.7604 & 0.5941 & 0.8635 & 0.7652 & 0.6430 & 0.9258 & 0.9117 & 0.9283 & 0.9461 & 0.8153 & 0.7252 & 0.9280 \\
\midrule
Mip-Splatting & 0.7669 & 0.6080 & 0.8689 & 0.7744 & 0.6367 & 0.9202 & 0.9097 & 0.9283 & 0.9427 & 0.8173 & 0.7310 & 0.9252 \\
Analytic-Splatting & 0.7620 & 0.6057 & 0.8659 & 0.7717 & 0.6325 & 0.9205 & 0.9094 & 0.9277 & 0.9439 & 0.8155 & 0.7276 & 0.9254 \\
AbsGS & 0.7830 & 0.6230 & 0.8710 & 0.7800 & 0.6170 & 0.9250 & 0.9110 & 0.9293 & 0.9450 & 0.8205 & 0.7348 & 0.9276 \\
Mip-Splatting+AbsGS & \cellcolor[HTML]{FF9999}0.7923 & \cellcolor[HTML]{FFF6B2}0.6397 & \cellcolor[HTML]{FFF6B2}0.8764 & \cellcolor[HTML]{FF9999}0.7908 & 0.6360 & 0.9255 & 0.9135 & 0.9304 & 0.9465 & \cellcolor[HTML]{FFF6B2}0.8279 & \cellcolor[HTML]{FFCC99}0.7470 & 0.9290 \\
Analytic-Splatting+AbsGS & \cellcolor[HTML]{FFCC99}0.7920 & \cellcolor[HTML]{FFCC99}0.6398 & \cellcolor[HTML]{FF9999}0.8779 & \cellcolor[HTML]{FF9999}0.7908 & 0.6365 & 0.9257 & 0.9142 & 0.9304 & 0.9466 & \cellcolor[HTML]{FFCC99}0.8282 & \cellcolor[HTML]{FF9999}0.7474 & 0.9292 \\
\makecell[l]{\textbf{3DGabSplat}\\ \textbf{ +Mip-Splatting+AbsGS}} & \cellcolor[HTML]{FFF6B2}0.7897 & \cellcolor[HTML]{FF9999}0.6402 & \cellcolor[HTML]{FFCC99}0.8768 & \cellcolor[HTML]{FFF6B2}0.7890 & 0.6322 & \cellcolor[HTML]{FFCC99}0.9291 & \cellcolor[HTML]{FFCC99}0.9170 & \cellcolor[HTML]{FF9999}0.9329 & \cellcolor[HTML]{FFCC99}0.9492 & \cellcolor[HTML]{FF9999}0.8285 & \cellcolor[HTML]{FFF6B2}0.7456 & \cellcolor[HTML]{FFCC99}0.9321\\
\bottomrule
\end{tabular}
\vspace{6pt}
\setlength{\tabcolsep}{3.5pt}
\centering
\caption{Per-scene PSNR (dB) comparison on Mip-NeRF360 Dataset.}
\label{tab:mip360-psnr}
\begin{tabular}{@{}l ccccc cccc ccc@{}}
\toprule
\multirow{2}*{Method} & \multicolumn{5}{c}{Outdoor} & \multicolumn{4}{c}{Indoor} & \multirow{2}*{Average} & \multirow{2}*{Outdoor} & \multirow{2}*{Indoor} \\
\cmidrule(lr){2-6}\cmidrule(lr){7-10}
& bicycle & flowers & garden & stump & treehill & room & counter & kitchen & bonsai & & & \\ 
\midrule
Mip-NeRF360 & 24.3700 & \cellcolor[HTML]{FF9999}21.7300 & 26.9800 & 26.4000 & 22.8700 & 31.6300 & 29.5500 & \cellcolor[HTML]{FF9999}32.2300 & \cellcolor[HTML]{FF9999}33.4600 & 27.6911 & 24.4700 & \cellcolor[HTML]{FFF6B2}31.7175 \\
3DGS & 25.2500 & 21.5200 & 27.4100 & 26.5500 & 22.4900 & 30.6300 & 28.7000 & 30.3200 & 31.9800 & 27.2056 & 24.6440 & 30.4075 \\
3DGS* & 25.1461 & 21.4960 & 27.4257 & 26.5519 & 22.4008 & 31.4579 & 29.0153 & 31.0659 & 32.1138 & 27.4082 & 24.6041 & 30.9132 \\
2DGS & 24.8200 & 20.9900 & 26.9100 & 26.4100 & 22.5200 & 30.8600 & 28.4500 & 30.6200 & 31.6400 & 27.0244 & 24.3300 & 30.3925 \\
GES-densify\_grad\_th\_3e-4 & 24.7682 & 20.6789 & 27.0067 & 26.0300 & 22.2884 & 31.1167 & 28.6151 & 30.7332 & 31.5824 & 26.9800 & 24.1545 & 30.5118 \\
GES-densify\_grad\_th\_2e-4 & 25.1158 & 21.4288 & 27.2633 & 26.4734 & 22.3742 & 31.2690 & 28.8786 & 31.2513 & 32.0972 & 27.3502 & 24.5311 & 30.8740 \\
3DCS & 24.7200 & 20.5200 & 27.0900 & 26.1200 & 21.7700 & 31.7000 & 29.0200 & 31.9600 & 32.6400 & 27.2822 & 24.0440 & 31.3300 \\
\makecell[l]{\textbf{3DGaborSplat}\\ \textbf{(Proposed)}} & 25.2893 & \cellcolor[HTML]{FFF6B2}21.6571 & 27.6648 & 26.6425 & 22.3780 & \cellcolor[HTML]{FFCC99}32.2731 & 29.5670 & \cellcolor[HTML]{FFF6B2}32.0546 & \cellcolor[HTML]{FFF6B2}33.0762 & \cellcolor[HTML]{FFCC99}27.8447 & 24.7263 & \cellcolor[HTML]{FFCC99}31.7427 \\ 
\midrule
Scaffold-GS & 25.0200 & 21.2500 & 27.3200 & 26.6500 & \cellcolor[HTML]{FF9999}23.2100 & 31.9000 & \cellcolor[HTML]{FFF6B2}29.6100 & 31.7300 & 32.7800 & 27.7189 & 24.6900 & 31.5050 \\
Scaffold-GS* & 25.1790 & 21.4482 & 27.5295 & 26.5596 & 22.9912 & 32.0772 & 29.5211 & 31.6099 & 32.5811 & 27.7219 & 24.7415 & 31.4473 \\
Octree-GS & 25.0500 & 21.2300 & 27.5100 & 26.4600 & \cellcolor[HTML]{FFF6B2}23.0000 & 32.1000 & \cellcolor[HTML]{FF9999}29.8800 & 31.3000 & 33.0400 & 27.7300 & 24.6500 & 31.5800 \\
Octree-GS* & 25.1227 & 21.4312 & 27.6449 & 26.4435 & 22.7982 & 32.1015 & 29.4248 & 31.0052 & 31.4138 & 27.4873 & 24.6881 & 30.9863 \\
\makecell[l]{\textbf{3DGaborSplat}\\ \textbf{ +Scaffold-GS}} & 25.2363 & 21.4445 & 27.5303 & 26.6036 & \cellcolor[HTML]{FFCC99}23.1059 & \cellcolor[HTML]{FF9999}32.2947 & 29.5050 & 31.7803 & 33.0512 & \cellcolor[HTML]{FFF6B2}27.8391 & \cellcolor[HTML]{FFF6B2}24.7841 & 31.6578 \\ 
\midrule
Mip-Splatting & 25.3110 & 21.6221 & 27.4532 & 26.6218 & 22.6282 & 31.6220 & 29.1172 & 31.5359 & 32.3058 & 27.5797 & 24.7273 & 31.1452 \\
Analytic-Splatting & 25.1822 & 21.6054 & 27.3910 & 26.6448 & 22.5381 & 31.7448 & 29.1039 & 31.5577 & 32.4320 & 27.5778 & 24.6723 & 31.2096 \\
AbsGS & 25.2900 & 21.3470 & 27.4870 & 26.7110 & 21.9860 & 31.6140 & 29.0310 & 31.6210 & 32.3230 & 27.4900 & 24.5642 & 31.1473 \\
Mip-Splatting+AbsGS & \cellcolor[HTML]{FFF6B2}25.5003 & \cellcolor[HTML]{FFCC99}21.6658 & \cellcolor[HTML]{FFF6B2}27.6650 & \cellcolor[HTML]{FFF6B2}26.8822 & 22.1586 & 31.5532 & 29.2312 & 31.7339 & 32.5702 & 27.6623 & 24.7744 & 31.2721 \\
Analytic-Splatting+AbsGS & \cellcolor[HTML]{FFCC99}25.5462 & 21.5940 & \cellcolor[HTML]{FFCC99}27.7401 & \cellcolor[HTML]{FFCC99}26.9736 & 22.2343 & 31.5638 & 29.2687 & 31.6645 & 32.3662 & 27.6613 & \cellcolor[HTML]{FFCC99}24.8176 & 31.2158 \\
\makecell[l]{\textbf{3DGabSplat}\\ \textbf{ +Mip-Splatting+AbsGS}} & \cellcolor[HTML]{FF9999}25.5500 & 21.6410 & \cellcolor[HTML]{FF9999}27.8123 & \cellcolor[HTML]{FF9999}26.9845 & 22.1305 & \cellcolor[HTML]{FFF6B2}32.1363 & \cellcolor[HTML]{FFCC99}29.6630 & \cellcolor[HTML]{FFCC99}32.1988 & \cellcolor[HTML]{FFCC99}33.2301 & \cellcolor[HTML]{FF9999}27.9274 & \cellcolor[HTML]{FF9999}24.8237 & \cellcolor[HTML]{FF9999}31.8071\\
\bottomrule
\end{tabular}
\end{table*}
\begin{table*}[!p]
\setlength{\tabcolsep}{4pt}
\centering
\caption{Per-scene LPIPS comparison on Mip-NeRF360 Dataset.}
\label{tab:mip360-lpips}
\begin{tabular}{@{}l ccccc cccc ccc@{}}
\toprule
\multirow{2}*{Method} & \multicolumn{5}{c}{Outdoor} & \multicolumn{4}{c}{Indoor} & \multirow{2}*{Average} & \multirow{2}*{Outdoor} & \multirow{2}*{Indoor} \\
\cmidrule(lr){2-6}\cmidrule(lr){7-10}
& bicycle & flowers & garden & stump & treehill & room & counter & kitchen & bonsai & & & \\ 
\midrule
Mip-NeRF360 & 0.3010 & 0.3440 & 0.1700 & 0.2610 & 0.3390 & 0.2110 & 0.2040 & 0.1270 & \cellcolor[HTML]{FFCC99}0.1760 & 0.2370 & 0.2830 & 0.1795 \\
3DGS & 0.2050 & 0.3360 & 0.1030 & 0.2100 & 0.3170 & 0.2200 & 0.2040 & 0.1290 & 0.2050 & 0.2143 & 0.2342 & 0.1895 \\
3DGS* & 0.2106 & 0.3352 & 0.1067 & 0.2161 & 0.3253 & 0.2185 & 0.1998 & 0.1267 & 0.2035 & 0.2158 & 0.2388 & 0.1871 \\
2DGS & 0.2710 & 0.3780 & 0.1380 & 0.2630 & 0.3690 & 0.2140 & 0.1970 & 0.1250 & 0.1940 & 0.2388 & 0.2838 & 0.1825 \\
GES-densify\_grad\_th\_3e-4 & 0.2684 & 0.3959 & 0.1377 & 0.2635 & 0.3789 & 0.2349 & 0.2147 & 0.1364 & 0.2168 & 0.2497 & 0.2889 & 0.2007 \\
GES-densify\_grad\_th\_2e-4 & 0.2120 & 0.3386 & 0.1086 & 0.2174 & 0.3278 & 0.2190 & 0.2005 & 0.1271 & 0.2044 & 0.2173 & 0.2409 & 0.1877 \\
3DCS & 0.2160 & 0.3220 & 0.1130 & 0.2270 & 0.3170 & \cellcolor[HTML]{FFF6B2}0.1930 & \cellcolor[HTML]{FFCC99}0.1820 & \cellcolor[HTML]{FFCC99}0.1170 & 0.1820 & 0.2077 & 0.2390 & \cellcolor[HTML]{FFF6B2}0.1685 \\
\makecell[l]{\textbf{3DGabSplat}\\ \textbf{(Proposed)}} & 0.2031 & 0.3299 & 0.1036 & 0.2140 & 0.3174 & 0.2111 & 0.1936 & 0.1224 & 0.1975 & 0.2103 & 0.2336 & 0.1811 \\
\midrule
Scaffold-GS & 0.2660 & 0.3720 & 0.1340 & 0.2600 & 0.3470 & \cellcolor[HTML]{FFCC99}0.1910 & 0.1840 & \cellcolor[HTML]{FFCC99}0.1170 & \cellcolor[HTML]{FFF6B2}0.1780 & 0.2277 & 0.2758 & \cellcolor[HTML]{FFCC99}0.1675 \\
Scaffold-GS* & 0.2270 & 0.3469 & 0.1179 & 0.2345 & 0.3199 & 0.2094 & 0.1985 & 0.1271 & 0.2025 & 0.2204 & 0.2492 & 0.1844 \\
Octree-GS & 0.2480 & 0.3550 & 0.1190 & 0.2650 & 0.3350 & \cellcolor[HTML]{FF9999}0.1780 & \cellcolor[HTML]{FF9999}0.1690 & \cellcolor[HTML]{FF9999}0.1140 & \cellcolor[HTML]{FF9999}0.1670 & 0.2167 & 0.2644 & \cellcolor[HTML]{FF9999}0.1570 \\
Octree-GS* & 0.2192 & 0.3377 & 0.1100 & 0.2361 & 0.3028 & 0.2061 & 0.2027 & 0.1363 & 0.2199 & 0.2190 & 0.2412 & 0.1912 \\
\makecell[l]{\textbf{3DGabSplat}\\ \textbf{ +Scaffold-GS}} & 0.2261 & 0.3474 & 0.1165 & 0.2365 & 0.3203 & 0.2080 & 0.1993 & 0.1265 & 0.2013 & 0.2202 & 0.2494 & 0.1838 \\ \midrule
Mip-Splatting & 0.2127 & 0.3400 & 0.1081 & 0.2161 & 0.3291 & 0.2213 & 0.2010 & 0.1266 & 0.2078 & 0.2181 & 0.2412 & 0.1892 \\
Analytic-Splatting & 0.2128 & 0.3358 & 0.1089 & 0.2181 & 0.3266 & 0.2206 & 0.2008 & 0.1271 & 0.2060 & 0.2174 & 0.2404 & 0.1886 \\
AbsGS & 0.1710 & \cellcolor[HTML]{FFCC99}0.2700 & 0.1000 & 0.1950 & 0.2780 & 0.2000 & 0.1890 & 0.1210 & 0.1900 & 0.1904 & 0.2028 & 0.1750 \\
Mip-Splatting+AbsGS & \cellcolor[HTML]{FFCC99}0.1671 & \cellcolor[HTML]{FFF6B2}0.2714 & \cellcolor[HTML]{FFF6B2}0.0942 & \cellcolor[HTML]{FFCC99}0.1887 & \cellcolor[HTML]{FFCC99}0.2747 & 0.2009 & 0.1869 & 0.1198 & 0.1869 & \cellcolor[HTML]{FFCC99}0.1878 & \cellcolor[HTML]{FFCC99}0.1992 & 0.1736 \\
Analytic-Splatting+AbsGS & \cellcolor[HTML]{FFF6B2}0.1678 & 0.2731 & \cellcolor[HTML]{FFCC99}0.0941 & \cellcolor[HTML]{FFF6B2}0.1896 & \cellcolor[HTML]{FFF6B2}0.2770 & 0.2046 & 0.1882 & 0.1211 & 0.1922 & \cellcolor[HTML]{FFF6B2}0.1898 & \cellcolor[HTML]{FFF6B2}0.2003 & 0.1765 \\
\makecell[l]{\textbf{3DGabSplat}\\ \textbf{ +Mip-Splatting+AbsGS}} & \cellcolor[HTML]{FF9999}0.1669 & \cellcolor[HTML]{FF9999}0.2646 & \cellcolor[HTML]{FF9999}0.0930 & \cellcolor[HTML]{FF9999}0.1880 & \cellcolor[HTML]{FF9999}0.2721 & 0.1976 & \cellcolor[HTML]{FFF6B2}0.1837 & 0.1172 & 0.1839 & \cellcolor[HTML]{FF9999}0.1852 & \cellcolor[HTML]{FF9999}0.1969 & 0.1706
\\
\bottomrule
\end{tabular}
\setlength{\tabcolsep}{10pt}
\centering
\caption{Per-scene SSIM comparison on the Tanks\&Temples (T\&T) and Deep Blending (DB) Datasets.}
\label{tab:tnt-ssim}
\begin{tabular}{@{}l cc cc ccc@{}}
\toprule
\multirow{2}*{Method} & \multicolumn{2}{c}{Tanks\&Temples} & \multicolumn{2}{c}{Deep Blending} & \multirow{2}*{Average} & \multirow{2}*{T\&T} & \multirow{2}*{DB} \\
\cmidrule(lr){2-3}\cmidrule(lr){4-5}
& truck & train & drjohnson & playroom & & & \\ 
\midrule
Mip-NeRF360 & 0.8570 & 0.6600 & 0.9010 & 0.9000 & 0.8295 & 0.7585 & 0.9005 \\
3DGS & 0.8790 & 0.8020 & 0.8990 & 0.9060 & 0.8715 & 0.8405 & 0.9025 \\
3DGS* & 0.8821 & 0.8138 & 0.9000 & 0.9071 & 0.8758 & 0.8479 & 0.9036 \\
2DGS & 0.8732 & 0.7908 & 0.8992 & 0.9063 & 0.8674 & 0.8320 & 0.9028 \\
GES-densify\_grad\_th\_3e-4 & 0.8774 & 0.8021 & 0.9024 & 0.9085 & 0.8726 & 0.8398 & 0.9055 \\
GES-densify\_grad\_th\_2e-4 & 0.8827 & 0.8168 & 0.9017 & 0.9082 & 0.8774 & 0.8498 & 0.9050 \\
3DCS & 0.8820 & 0.8200 & 0.9020 & 0.9020 & 0.8765 & 0.8510 & 0.9020 \\
\textbf{3DGabSplat (Proposed)} & 0.8851 & 0.8256 & 0.9068 & 0.9110 & 0.8821 & 0.8554 & 0.9089 \\
\midrule
Scaffold-GS & 0.8860 & 0.8210 & \cellcolor[HTML]{FFF6B2}0.9070 & \cellcolor[HTML]{FFF6B2}0.9130 & 0.8818 & 0.8535 & \cellcolor[HTML]{FFF6B2}0.9100 \\
Scaffold-GS* & 0.8859 & 0.8188 & 0.9064 & 0.9117 & 0.8807 & 0.8524 & 0.9091 \\
Octree-GS & \cellcolor[HTML]{FF9999}0.8960 & \cellcolor[HTML]{FF9999}0.8350 & \cellcolor[HTML]{FF9999}0.9120 & \cellcolor[HTML]{FF9999}0.9140 & \cellcolor[HTML]{FF9999}0.8893 & \cellcolor[HTML]{FF9999}0.8655 & \cellcolor[HTML]{FF9999}0.9130 \\
Octree-GS* & 0.8875 & \cellcolor[HTML]{FFCC99}0.8331 & 0.9043 & 0.9123 & 0.8843 & \cellcolor[HTML]{FFF6B2}0.8603 & 0.9083 \\
\textbf{3DGabSplat+Scaffold-GS} & 0.8874 & 0.8314 & \cellcolor[HTML]{FFCC99}0.9079 & \cellcolor[HTML]{FFCC99}0.9131 & \cellcolor[HTML]{FFCC99}0.8850 & 0.8594 & \cellcolor[HTML]{FFCC99}0.9105 \\
\midrule
Mip-Splatting & 0.8840 & 0.8182 & 0.9008 & 0.9085 & 0.8779 & 0.8511 & 0.9047 \\
Analytic-Splatting & 0.8831 & 0.8186 & 0.9014 & 0.9086 & 0.8779 & 0.8509 & 0.9050 \\
AbsGS & 0.8880 & 0.8180 & 0.8980 & 0.9070 & 0.8778 & 0.8530 & 0.9025 \\
Mip-Splatting+AbsGS & \cellcolor[HTML]{FFF6B2}0.8924 & 0.8248 & 0.8976 & 0.9073 & 0.8806 & 0.8586 & 0.9025 \\
Analytic-Splatting+AbsGS & \cellcolor[HTML]{FFF6B2}0.8924 & 0.8264 & 0.8992 & 0.9083 & 0.8816 & 0.8594 & 0.9038 \\
\textbf{3DGabSplat+Mip-Splatting+AbsGS} & \cellcolor[HTML]{FFCC99}0.8935 & \cellcolor[HTML]{FFF6B2}0.8315 & 0.9052 & 0.9090 & \cellcolor[HTML]{FFF6B2}0.8848 & \cellcolor[HTML]{FFCC99}0.8625 & 0.9071\\
\bottomrule
\end{tabular}
\end{table*}
\begin{table*}[!p]
\setlength{\tabcolsep}{9pt}
\centering
\caption{Per-scene PSNR (dB) comparison on the Tanks\&Temples (T\&T) and Deep Blending (DB) Datasets.}
\label{tab:tnt-psnr}
\begin{tabular}{@{}l cc cc ccc@{}}
\toprule
\multirow{2}*{Method} & \multicolumn{2}{c}{Tanks\&Temples} & \multicolumn{2}{c}{Deep Blending} & \multirow{2}*{Average} & \multirow{2}*{T\&T} & \multirow{2}*{DB} \\
\cmidrule(lr){2-3}\cmidrule(lr){4-5}
& truck & train & drjohnson & playroom & & & \\ 
\midrule
Mip-NeRF360 & 24.9120 & 19.5230 & 29.1400 & 29.6570 & 25.8080 & 22.2175 & 29.3985 \\
3DGS & 25.1870 & 21.0970 & 28.7660 & 30.0440 & 26.2735 & 23.1420 & 29.4050 \\
3DGS* & 25.4067 & 21.9817 & 29.0785 & 30.0220 & 26.6222 & 23.6942 & 29.5502 \\
2DGS & 25.0997 & 21.2134 & 28.8494 & 30.1483 & 26.3277 & 23.1566 & 29.4988 \\
GES-densify\_grad\_th\_3e-4 & 25.1454 & 21.6525 & 29.3056 & 30.1605 & 26.5660 & 23.3989 & 29.7331 \\
GES-densify\_grad\_th\_2e-4 & 25.2688 & 21.9677 & 29.1334 & 30.0496 & 26.6049 & 23.6182 & 29.5915 \\
3DCS & 25.6500 & 22.2300 & 29.5400 & 30.0800 & 26.8750 & 23.9400 & 29.8100 \\
\textbf{3DGabSplat (Proposed)} & 25.8756 & \cellcolor[HTML]{FF9999}23.1018 & 29.6138 & 30.5647 & \cellcolor[HTML]{FFF6B2}27.2890 & \cellcolor[HTML]{FFF6B2}24.4887 & 30.0893 \\ \midrule
Scaffold-GS & 25.8900 & 22.2000 & \cellcolor[HTML]{FFF6B2}29.7900 & \cellcolor[HTML]{FFCC99}31.0700 & 27.2375 & 24.0450 & \cellcolor[HTML]{FFCC99}30.4300 \\
Scaffold-GS* & 25.8025 & 22.1879 & 29.7590 & 30.8818 & 27.1578 & 23.9952 & 30.3204 \\
Octree-GS & \cellcolor[HTML]{FF9999}26.2700 & \cellcolor[HTML]{FFF6B2}22.7700 & \cellcolor[HTML]{FFCC99}29.8700 & \cellcolor[HTML]{FFF6B2}30.9500 & \cellcolor[HTML]{FFCC99}27.4650 & \cellcolor[HTML]{FF9999}24.5200 & \cellcolor[HTML]{FFF6B2}30.4100 \\
Octree-GS* & 25.9381 & 22.6648 & 29.5096 & 30.8266 & 27.2348 & 24.3015 & 30.1681 \\
\textbf{3DGabSplat+Scaffold-GS} & \cellcolor[HTML]{FFF6B2}26.0530 & \cellcolor[HTML]{FFCC99}22.9861 & \cellcolor[HTML]{FF9999}29.8764 & \cellcolor[HTML]{FF9999}31.1542 & \cellcolor[HTML]{FF9999}27.5174 & \cellcolor[HTML]{FFCC99}24.5196 & \cellcolor[HTML]{FF9999}30.5153 \\ \midrule
Mip-Splatting & 25.4814 & 22.0928 & 29.0913 & 30.2821 & 26.7369 & 23.7871 & 29.6867 \\
Analytic-Splatting & 25.4777 & 22.2002 & 29.1840 & 30.3191 & 26.7953 & 23.8390 & 29.7516 \\
AbsGS & 25.7350 & 21.7210 & 29.1970 & 30.1410 & 26.6985 & 23.7280 & 29.6690 \\
Mip-Splatting+AbsGS & 25.5957 & 21.8340 & 28.6922 & 29.9191 & 26.5102 & 23.7148 & 29.3057 \\
Analytic-Splatting+AbsGS & 25.6173 & 21.8128 & 28.8686 & 30.2618 & 26.6401 & 23.7151 & 29.5652 \\
\textbf{3DGabSplat+Mip-Splatting+AbsGS} & \cellcolor[HTML]{FFCC99}26.0589 & 22.6719 & 29.3067 & 30.2662 & 27.0759 & 24.3654 & 29.7865\\
\bottomrule
\end{tabular}
\vspace{12pt}
\centering
\caption{Per-scene LPIPS comparison on the Tanks\&Temples (T\&T) and Deep Blending (DB) Datasets.}
\label{tab:tnt-lpips}
\begin{tabular}{@{}l cc cc ccc@{}}
\toprule
\multirow{2}*{Method} & \multicolumn{2}{c}{Tanks\&Temples} & \multicolumn{2}{c}{Deep Blending} & \multirow{2}*{Average} & \multirow{2}*{T\&T} & \multirow{2}*{DB} \\
\cmidrule(lr){2-3}\cmidrule(lr){4-5}
& truck & train & drjohnson & playroom & & & \\ 
\midrule
Mip-NeRF360 & 0.1590 & 0.3540 & 0.2370 & 0.2520 & 0.2505 & 0.2565 & 0.2445 \\
3DGS & 0.1480 & 0.2180 & 0.2440 & 0.2410 & 0.2128 & 0.1830 & 0.2425 \\
3DGS* & 0.1465 & 0.2057 & 0.2443 & 0.2432 & 0.2099 & 0.1761 & 0.2437 \\
2DGS & 0.1728 & 0.2503 & 0.2570 & 0.2573 & 0.2343 & 0.2115 & 0.2571 \\
GES-densify\_grad\_th\_3e-4 & 0.1592 & 0.2319 & 0.2471 & 0.2507 & 0.2222 & 0.1956 & 0.2489 \\
GES-densify\_grad\_th\_2e-4 & 0.1460 & 0.2056 & 0.2428 & 0.2431 & 0.2094 & 0.1758 & 0.2429 \\
3DCS & 0.1250 & \cellcolor[HTML]{FFCC99}0.1870 & 0.2380 & 0.2370 & \cellcolor[HTML]{FFF6B2}0.1968 & \cellcolor[HTML]{FFF6B2}0.1560 & \cellcolor[HTML]{FFF6B2}0.2375 \\
\textbf{3DGabSplat (Proposed)} & 0.1430 & 0.1969 & \cellcolor[HTML]{FFF6B2}0.2349 & 0.2402 & 0.2037 & 0.1699 & \cellcolor[HTML]{FFF6B2}0.2375 \\ \midrule
Scaffold-GS & 0.1410 & 0.2040 & 0.2500 & 0.2500 & 0.2113 & 0.1725 & 0.2500 \\
Scaffold-GS* & 0.1422 & 0.2079 & 0.2526 & 0.2530 & 0.2139 & 0.1751 & 0.2528 \\
Octree-GS & \cellcolor[HTML]{FF9999}0.1180 & 0.1880 & \cellcolor[HTML]{FF9999}0.2310 & 0.2440 & \cellcolor[HTML]{FFCC99}0.1953 & \cellcolor[HTML]{FF9999}0.1530 & \cellcolor[HTML]{FFF6B2}0.2375 \\
Octree-GS* & 0.1322 & \cellcolor[HTML]{FFF6B2}0.1877 & 0.2519 & 0.2502 & 0.2055 & 0.1599 & 0.2510 \\
\textbf{3DGabSplat+Scaffold-GS} & 0.1374 & 0.1945 & 0.2521 & 0.2534 & 0.2094 & 0.1660 & 0.2528 \\ \midrule
Mip-Splatting & 0.1486 & 0.2075 & 0.2475 & 0.2482 & 0.2130 & 0.1780 & 0.2479 \\
Analytic-Splatting & 0.1483 & 0.2065 & 0.2474 & 0.2482 & 0.2126 & 0.1774 & 0.2478 \\
AbsGS & 0.1310 & 0.1930 & 0.2400 & \cellcolor[HTML]{FFCC99}0.2320 & 0.1990 & 0.1620 & \cellcolor[HTML]{FFCC99}0.2360 \\
Mip-Splatting+AbsGS & \cellcolor[HTML]{FFF6B2}0.1238 & 0.1896 & 0.2432 & \cellcolor[HTML]{FFF6B2}0.2347 & 0.1978 & 0.1567 & 0.2389 \\
Analytic-Splatting+AbsGS & 0.1251 & 0.1908 & 0.2461 & 0.2396 & 0.2004 & 0.1580 & 0.2428 \\
\textbf{3DGabSplat+Mip-Splatting+AbsGS} & \cellcolor[HTML]{FFCC99}0.1235 & \cellcolor[HTML]{FF9999}0.1837 & \cellcolor[HTML]{FFCC99}0.2322 & \cellcolor[HTML]{FF9999}0.2308 & \cellcolor[HTML]{FF9999}0.1925 & \cellcolor[HTML]{FFCC99}0.1536 & \cellcolor[HTML]{FF9999}0.2315\\
\bottomrule
\end{tabular}
\end{table*}

\section{Implementation Details}
In this section, we provide a detailed description of the parameter settings and training configurations used in our experiments.
First, our proposed 3D Gabor-based primitive extends the vanilla Gaussian kernel by incorporating two additional parameters: the frequency and the weighting coefficient of each Gabor kernel.
For initialization, the frequency and coefficient are set to 0.05 and 0.01, respectively, for the MipNeRF360 dataset, 0.00001 and 0.02 for the Tanks and Temples (T\&T) dataset, and 0.001 and 0.01 for all other datasets.
In addition, during our frequency-adaptive optimization process, the coefficients are periodically reset to 0.01 every 3000 epochs, while both the frequency and coefficient are reset to 0.001 and 0.01, respectively, after densification.
For opacity, we increase the threshold to 0.01 and reset the opacity to 0.02 after densification.
The default learning rates for the frequency and coefficient parameters are set to 0.02 and 0.01, respectively, with the coefficient learning rate further reduced to 0.005 for the T\&T and Deep Blending (DB) datasets to ensure stable convergence.
The learning rate for opacity is set to 0.025 on the MipNeRF360 dataset, while it is adjusted to 0.01 and 0.02 for the T\&T and DB datasets, respectively.
For the MipNeRF360 dataset, outdoor and indoor scenes are trained using ground truth images at resolutions of 1/4 and 1/2 of the original, respectively. It is important to note that we use the downsampled images provided by the official source, rather than downsampling from the original resolution. The latter approach, while potentially yielding superior performance, would undermine the fairness of the comparison.
For the remaining datasets, including T\&T, DB, and NeRF Synthetic, we conduct supervised training using the original resolution.
During training, we follow the same configuration as 3DGS, setting the total number of iterations to 30k, with 15k iterations for densification.
All other parameters and configurations are kept consistent with 3DGS to ensure a fair comparison.

\begin{table*}[!t]
\setlength{\abovecaptionskip}{0pt}
\centering
\caption{Per-scene SSIM comparison on NeRF Synthetic Dataset.}
\label{tab:nerf-ssim}
\begin{tabular}{@{}l cccccccc c@{}}
\toprule
Method & chair & drums & ficus & hotdog & lego & materials & mic & ship & Average \\ 
\midrule
3DGS & \cellcolor[HTML]{FFCC99}0.9874 & \cellcolor[HTML]{FFCC99}0.9545 & 0.9871 & \cellcolor[HTML]{FFCC99}0.9852 & \cellcolor[HTML]{FFCC99}0.9830 & 0.9604 & 0.9914 & \cellcolor[HTML]{FFCC99}0.9065 & \cellcolor[HTML]{FFCC99}0.9694 \\
2DGS & 0.9849 & 0.9539 & \cellcolor[HTML]{FF9999}0.9880 & 0.9840 & 0.9802 & 0.9579 & 0.9904 & 0.9042 & 0.9679 \\
GES-densify\_grad\_th\_3e-4 & 0.9850 & 0.9535 & 0.9871 & 0.9840 & 0.9793 & 0.9584 & 0.9910 & 0.9033 & 0.9677 \\
GES-densify\_grad\_th\_2e-4 & \cellcolor[HTML]{FFF6B2}0.9873 & \cellcolor[HTML]{FFCC99}0.9545 & \cellcolor[HTML]{FFF6B2}0.9872 & \cellcolor[HTML]{FFCC99}0.9852 & \cellcolor[HTML]{FFF6B2}0.9825 & 0.9602 & 0.9914 & \cellcolor[HTML]{FFF6B2}0.9063 & \cellcolor[HTML]{FFF6B2}0.9693 \\
3DCS & 0.9738 & 0.9472 & 0.9846 & 0.9784 & 0.9671 & 0.9518 & 0.9815 & 0.8623 & 0.9558 \\
Scaffold-GS & 0.9846 & 0.9507 & 0.9861 & 0.9839 & 0.9800 & \cellcolor[HTML]{FFF6B2}0.9622 & \cellcolor[HTML]{FFCC99}0.9921 & 0.8992 & 0.9674 \\
Octree-GS & 0.9849 & 0.9485 & 0.9837 & 0.9841 & 0.9797 & \cellcolor[HTML]{FF9999}0.9627 & \cellcolor[HTML]{FFF6B2}0.9918 & 0.8983 & 0.9667 \\
\midrule
\textbf{3DGabSplat (Proposed)} & \cellcolor[HTML]{FF9999}0.9879 & \cellcolor[HTML]{FF9999}0.9553 & \cellcolor[HTML]{FFCC99}0.9873 & \cellcolor[HTML]{FF9999}0.9858 & \cellcolor[HTML]{FF9999}0.9836 & \cellcolor[HTML]{FFCC99}0.9626 & \cellcolor[HTML]{FF9999}0.9927 & \cellcolor[HTML]{FF9999}0.9081 & \cellcolor[HTML]{FF9999}0.9704\\
\bottomrule
\end{tabular}
\vspace{6pt}
\centering
\caption{Per-scene PSNR (dB) comparison on NeRF Synthetic Dataset.}
\label{tab:nerf-psnr}
\begin{tabular}{@{}l cccccccc c@{}}
\toprule
Method & chair & drums & ficus & hotdog & lego & materials & mic & ship & Average \\ 
\midrule
3DGS & \cellcolor[HTML]{FFCC99}35.8730 & \cellcolor[HTML]{FFF6B2}26.1411 & \cellcolor[HTML]{FFF6B2}34.8385 & \cellcolor[HTML]{FFCC99}37.7147 & \cellcolor[HTML]{FFCC99}35.8508 & 29.9635 & 35.3909 & \cellcolor[HTML]{FFCC99}30.9997 & \cellcolor[HTML]{FFCC99}33.3465 \\
2DGS & 35.2255 & 26.0903 & \cellcolor[HTML]{FF9999}35.3787 & 37.3685 & 35.1286 & 29.6730 & 34.9787 & 30.6299 & 33.0591 \\
GES-densify\_grad\_th\_3e-4 & 35.2198 & 26.0545 & 34.7472 & 37.3897 & 34.8389 & 29.5994 & 35.1789 & 30.6105 & 32.9549 \\
GES-densify\_grad\_th\_2e-4 & \cellcolor[HTML]{FFF6B2}35.8093 & 26.1298 & 34.7831 & 37.5927 & \cellcolor[HTML]{FFF6B2}35.6300 & 29.9080 & 35.2968 & \cellcolor[HTML]{FFF6B2}30.8342 & \cellcolor[HTML]{FFF6B2}33.2480 \\
3DCS & 32.9718 & 25.6063 & 34.4114 & 36.0833 & 32.6263 & 28.9986 & 32.1020 & 26.2702 & 31.1337 \\
Scaffold-GS & 35.1905 & \cellcolor[HTML]{FFCC99}26.3423 & 34.4868 & \cellcolor[HTML]{FFF6B2}37.6151 & 34.8931 & \cellcolor[HTML]{FFF6B2}30.2175 & \cellcolor[HTML]{FFCC99}36.2950 & 30.0388 & 33.1349 \\
Octree-GS & 35.2354 & 26.0780 & 33.5539 & 37.5500 & 34.8776 & \cellcolor[HTML]{FFCC99}30.2833 & \cellcolor[HTML]{FFF6B2}36.0209 & 30.0136 & 32.9516 \\ \midrule
\textbf{3DGabSplat (Proposed)} & \cellcolor[HTML]{FF9999}36.0774 & \cellcolor[HTML]{FF9999}26.4029 & \cellcolor[HTML]{FFCC99}35.1496 & \cellcolor[HTML]{FF9999}37.9374 & \cellcolor[HTML]{FF9999}36.2376 & \cellcolor[HTML]{FF9999}30.3332 & \cellcolor[HTML]{FF9999}36.3160 & \cellcolor[HTML]{FF9999}31.3239 & \cellcolor[HTML]{FF9999}33.7223
\\
\bottomrule
\end{tabular}
\vspace{6pt}
\centering
\caption{Per-scene LPIPS comparison on NeRF Synthetic Dataset.}
\label{tab:syn-lpips}
\begin{tabular}{@{}l cccccccc c@{}}
\toprule
Method & chair & drums & ficus & hotdog & lego & materials & mic & ship & Average \\ 
\midrule
3DGS & \cellcolor[HTML]{FFCC99}0.0117 & \cellcolor[HTML]{FFCC99}0.0368 & \cellcolor[HTML]{FF9999}0.0117 & \cellcolor[HTML]{FFCC99}0.0201 & \cellcolor[HTML]{FFCC99}0.0153 & \cellcolor[HTML]{FFCC99}0.0336 & \cellcolor[HTML]{FFF6B2}0.0062 & \cellcolor[HTML]{FFCC99}0.1065 & \cellcolor[HTML]{FFCC99}0.0302 \\
2DGS & 0.0147 & 0.0397 & \cellcolor[HTML]{FF9999}0.0117 & 0.0226 & 0.0195 & 0.0385 & 0.0071 & 0.1098 & 0.0330 \\
GES-densify\_grad\_th\_3e-4 & 0.0157 & 0.0395 & 0.0119 & 0.0237 & 0.0210 & 0.0369 & 0.0066 & 0.1148 & 0.0338 \\
GES-densify\_grad\_th\_2e-4 & \cellcolor[HTML]{FFF6B2}0.0119 & \cellcolor[HTML]{FFF6B2}0.0369 & \cellcolor[HTML]{FF9999}0.0117 & \cellcolor[HTML]{FFCC99}0.0201 & \cellcolor[HTML]{FFF6B2}0.0160 & \cellcolor[HTML]{FFF6B2}0.0343 & \cellcolor[HTML]{FFCC99}0.0061 & \cellcolor[HTML]{FFF6B2}0.1066 & \cellcolor[HTML]{FFF6B2}0.0304 \\
3DCS & 0.0291 & 0.0486 & 0.0164 & 0.0316 & 0.0364 & 0.0497 & 0.0168 & 0.1519 & 0.0476 \\
Scaffold-GS & 0.0145 & 0.0429 & 0.0126 & 0.0227 & 0.0189 & 0.0357 & 0.0066 & 0.1120 & 0.0332 \\
Octree-GS & 0.0144 & 0.0461 & 0.0151 & 0.0217 & 0.0192 & 0.0346 & 0.0066 & 0.1106 & 0.0335 \\ \midrule
\textbf{3DGabSplat (Proposed)} & \cellcolor[HTML]{FF9999}0.0115 & \cellcolor[HTML]{FF9999}0.0364 & \cellcolor[HTML]{FF9999}0.0117 & \cellcolor[HTML]{FF9999}0.0198 & \cellcolor[HTML]{FF9999}0.0151 & \cellcolor[HTML]{FF9999}0.0326 & \cellcolor[HTML]{FF9999}0.0054 & \cellcolor[HTML]{FF9999}0.1049 & \cellcolor[HTML]{FF9999}0.0297\\
\bottomrule
\end{tabular}
\end{table*}

\section{Additional Results}
We report an efficiency evaluation of our proposed 3DGabSplat against the baselines, as shown in Table \ref{tab:efficiency1}. To further validate our method in large, complex scenes, we conduct experiments on two additional datasets, two from OMMO [35] (only 01 and 03 scenes) and all eight scenes from BungeeNeRF [54], both featuring large-scale urban environments that present more complex scenes and greater reconstruction challenges. Results are shown in Table \ref{tab:efficiency2}. Our method outperforms baselines by 0.94dB (OMMO) and 0.27dB (BungeeNeRF) in PSNR on average, and uses fewer primitives and achieves higher rendering speed with similar memory usage.

To further demonstrate the effectiveness of our method, we compare it with recent state-of-the-art method 3DHGS [31] and DBS [34]. 3D-HGS splits Gaussians along a central plane with different opacities. DBS uses deformable Beta kernel with adjustable support. Both of them broaden frequency bandwidth but retain exponential decay, limiting band-pass modeling. In contrast, our method composes 3D Gabor-based primitives from kernels across multiple bands, enabling more effective band-pass filtering and better capturing high-frequency details. We compare with them under identical settings on MipNeRF360 in Table~\ref{tab:new_sota}. As DBS builds on 3DGS-MCMC [26], we use 3DGabSplat+MCMC for fairness. Our method outperforms both methods and achieves new SOTA with MCMC on this benchmark.

\begin{figure}[!t]
  \centering
  \includegraphics[width=\linewidth]{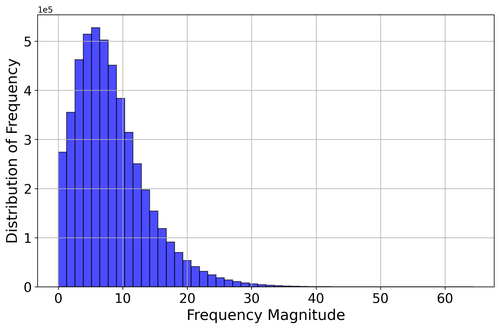}
  \caption{Visualization of the frequency amplitude distribution for each primitive in the bicycle scene.}
  \label{fig:freq_distribution}
\end{figure}

We also provide additional qualitative rendering comparisons, as shown in \figurename~\ref{fig:supple_visualization}. Experimental results demonstrate that our method achieves superior rendering quality in regions with complex textures and high-frequency details, such as the road surrounding the bicycle, flowers, and indoor scenes, whereas 3DGS and GES tend to produce blurry outputs and 3DCS often suffers from noticeable fragmentation.
The results further highlight that our 3DGabSplat, as a superior alternative to the Gaussian kernel, outperforms 3DGS and its variants utilizing different kernels. 
Moreover, the results clearly validate the effectiveness of our 3D Gabor-based primitive and the frequency-adaptive optimization strategy in faithfully capturing high-frequency details and intricate texture patterns in 3D scenes.
Per-scene quantitative comparisons are reported in Tables \ref{tab:mip360-ssim}--\ref{tab:syn-lpips}.

\begin{figure*}[!t]
\setlength{\abovecaptionskip}{5pt}
\centering
\begin{subfigure}{0.48\linewidth}
\centering
\includegraphics[width=\linewidth]{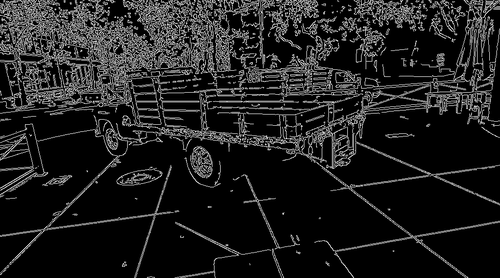}
\caption{Canny edge detection results.}
\end{subfigure}
\begin{subfigure}{0.48\linewidth}
\centering
\includegraphics[width=\linewidth]{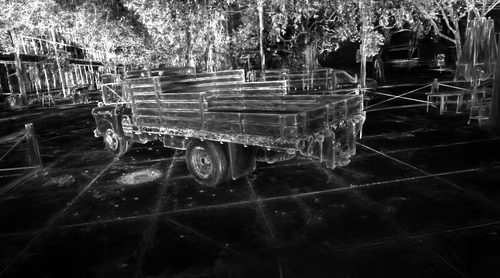}
\caption{Frequency Map.}
\end{subfigure}
\vspace{2mm}
\begin{subfigure}{0.48\linewidth}
\centering
\includegraphics[width=\linewidth]{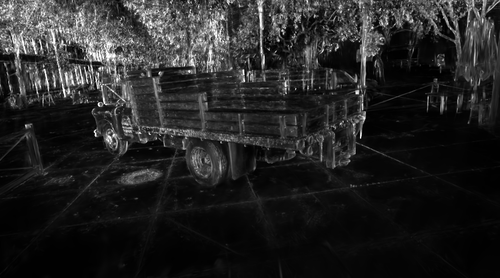}
\caption{X-axis Frequency Map}
\end{subfigure}
\begin{subfigure}{0.48\linewidth}
\centering
\includegraphics[width=\linewidth]{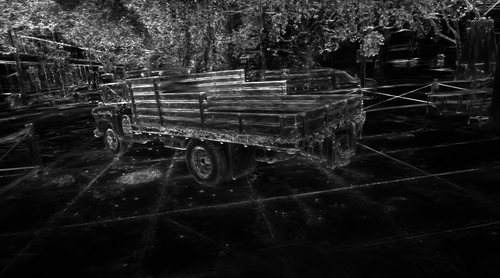} %
\caption{Y-axis Frequency Map} %
\end{subfigure}
\caption{Visualization of rendered frequency maps on the truck scene.}
\label{fig:freq_map}
\end{figure*}
\begin{figure*}[!t]
\setlength{\abovecaptionskip}{5pt}
\centering
\begin{subfigure}{0.32\linewidth}
\centering
\includegraphics[width=\linewidth]{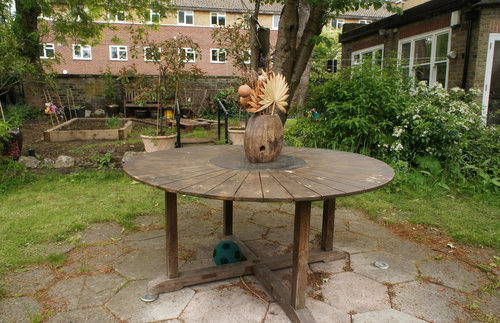}
\caption{Ground Truth}
\end{subfigure}
\begin{subfigure}{0.32\linewidth}
\centering
\includegraphics[width=\linewidth]{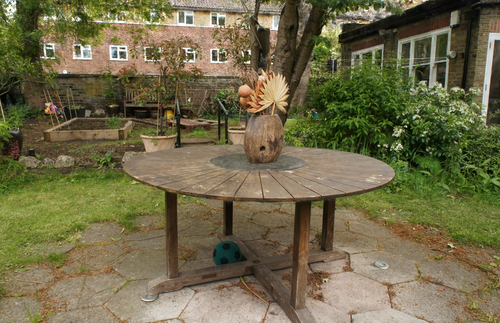}
\caption{Ours}
\end{subfigure}
\begin{subfigure}{0.32\linewidth}
\centering
\includegraphics[width=\linewidth]{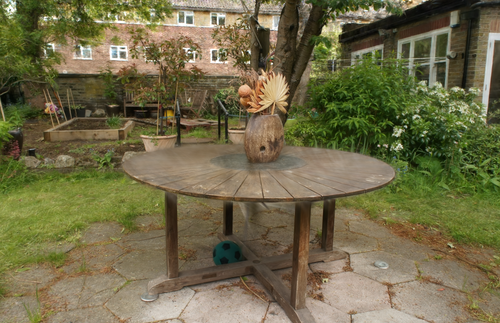}
\caption{Ours (degrade to Gaussian)}
\end{subfigure}
\caption{Visualization of degradation to Gaussian on the garden scene.}
\label{fig:degrade}
\end{figure*}

To further analyze the characteristics of Gabor-based primitives, we visualize the frequency amplitude distribution for each primitive in \figurename~\ref{fig:freq_distribution}. The results reveal a right-skewed distribution, where frequencies are predominantly concentrated in the low-frequency range and gradually diminish toward higher frequencies. We also visualize frequency maps by blending the weighted Gabor frequencies of each primitive. The high-magnitude regions in the frequency maps align well with the results of Canny edge detection, indicating that our Gabor-based primitives effectively capture fine textures and high-frequency details in the rendered images, highlighting the strength of Gabor kernels.
Directional frequency maps along the X- and Y-axes further highlight orientation-specific high-frequency content (\figurename~\ref{fig:freq_map}). Removing Gabor kernels results in noticeable blurring and over-smoothing in complex texture regions, along with significant PSNR degradation, further validating their role in preserving high-frequency details (\figurename~\ref{fig:degrade}).